\PassOptionsToPackage{table}{xcolor}

\documentclass[10pt,twocolumn,letterpaper]{article}

\usepackage[pagenumbers]{wacv} %

\usepackage{graphicx}
\usepackage{amsmath}
\usepackage{amssymb}
\usepackage{booktabs}
\usepackage{array}
\usepackage{xcolor,colortbl}
\usepackage{pgf}
\usepackage{booktabs}
\usepackage{array}
\usepackage{enumitem}
\usepackage{tabularx}
\usepackage{microtype}
\usepackage{flushend}

\definecolor{myblue}{RGB}{70, 130, 180} 

\newcommand{\colormap}[2]{%
  \colorbox{myblue!#1!white}{#2}%
}

\newcommand{\colorcell}[2]{%
  \pgfmathsetmacro{\diffpercent}{(#1 - #2)/#2 * 100}%
  \ifdim \diffpercent pt > 5pt
    \cellcolor{green!30}#1%
  \else
    \ifdim \diffpercent pt < -5pt
      \cellcolor{red!30}#1%
    \else
      \cellcolor{yellow!30}#1%
    \fi
  \fi
}

\usepackage[pagebackref,breaklinks,colorlinks]{hyperref}

\usepackage[capitalize]{cleveref}
\crefname{section}{Sec.}{Secs.}
\Crefname{section}{Section}{Sections}
\Crefname{table}{Table}{Tables}
\crefname{table}{Tab.}{Tabs.}

\def\wacvPaperID{719} %
\def\confName{WACV}
\def\confYear{2025}

\begin{document}

\title{Are All Marine Species Created Equal? \\ Performance Disparities in Underwater Object Detection}

\author{Melanie Wille \qquad \qquad Tobias Fischer \qquad \qquad Scarlett Raine\\[0.25cm]
Queensland University of Technology\\
Brisbane, Australia\\
{\tt\small \{melanie.wille,tobias.fischer,sg.raine\}@qut.edu.au}
}
\maketitle

\begin{abstract}
Underwater object detection is critical for monitoring marine ecosystems but poses unique challenges, including degraded image quality, imbalanced class distribution, and distinct visual characteristics. Not every species is detected equally well, yet underlying causes remain unclear. We address two key research questions: 1) What factors beyond data quantity drive class-specific performance disparities? 2) How can we systematically improve detection of under-performing marine species?
We manipulate the DUO and RUOD datasets to separate the object detection task into localization and classification and investigate the under-performance of the scallop class. Localization analysis using YOLO11 and TIDE finds that foreground-background discrimination is the most problematic stage regardless of data quantity. Classification experiments reveal persistent precision gaps even with balanced data, indicating intrinsic feature-based challenges beyond data scarcity and inter-class dependencies. We recommend imbalanced distributions when prioritizing precision, and balanced distributions when prioritizing recall. Improving under-performing classes should focus on algorithmic advances, especially within localization modules. We publicly release our code and datasets.
\end{abstract}

\vspace{-0.5cm}
\section{Introduction}
\label{sec:intro}
Oceans are crucial for the regulation and function of natural water and carbon cycles, making healthy marine ecosystems essential for life on Earth. However these ecosystems are under threat from anthropogenic impacts including pollution, over-fishing and climate change~\cite{raine2024reducing}. This increases the urgency and importance of monitoring these environments.  

The increasing use of Autonomous Underwater Vehicles (AUVs) and Remotely Operated Vehicles (ROVs) in recent years has yielded large quantities of underwater imagery~\cite{lucas2025underwater}. Object detection plays an important role in analyzing this data, by identifying and precisely locating specific marine targets within an image. This enables monitoring ecosystem health through estimates of species distribution, size of targets, population density, signs of diseases and the presence/absence of indicator species~\cite{er2023research}. 

\begin{figure}[t]
  \centering
  \includegraphics[width=\linewidth]{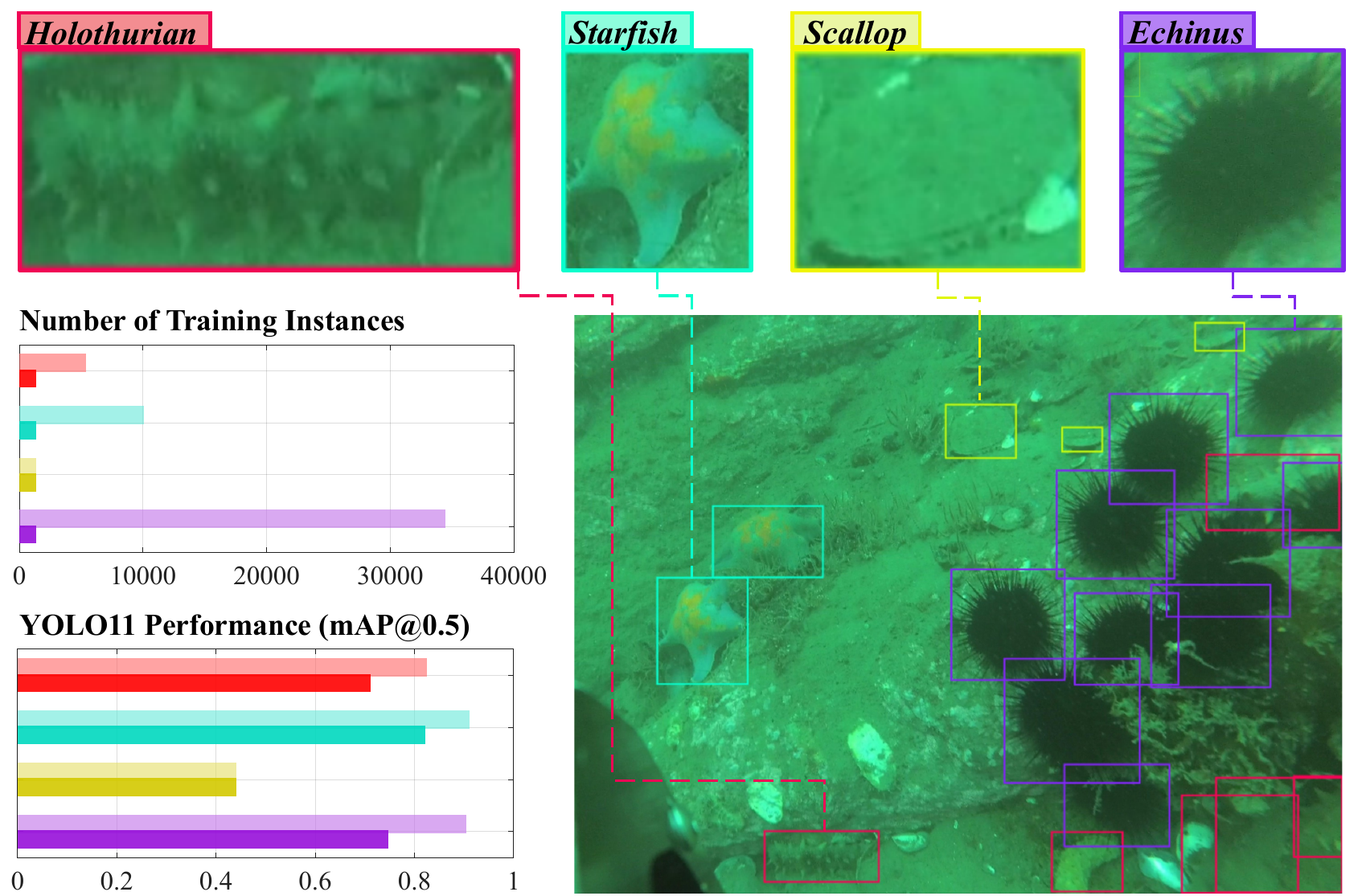}
  \caption{Example underwater image from the DUO dataset~\cite{liu2021dataset}, showing the four target classes (top row): \textit{Holothurian}, \textit{Starfish}, \textit{Scallop} and \textit{Echinus}. Left middle: the original class distribution of DUO shown by the lighter colored bars on top, and the DUO dataset with down-sampled, class-balanced distribution in the solid bars below. Left bottom: localization performance (mAP@0.5) on single-class versions of DUO. The lightened top bars indicate the localization performance for each class under the original distribution, and the solid colored bars below demonstrates there is still a clear performance difference between the classes, even when there is an equal number of training instances.}
  \label{fig:DUO_class-specific_performance}
  \vspace*{-0.3cm}
\end{figure}

However, object detection in underwater imagery is significantly more challenging than for terrestrial images. Not only is the image quality often degraded by turbidity, blur and low contrast~\cite{yuan2023multi}, but there are certain visual characteristics and behavioral habits associated with different marine species that complicate their reliable detection. Image datasets of marine species may exhibit class imbalance, varying sizes of individuals, intra-class variation in appearance, background camouflaging, clustering of individuals, and dense habitats. Naturally, this leads to a discrepancy in the object detection performance for different classes of marine species~(\cref{fig:DUO_class-specific_performance}).

The Detecting Underwater Objects (DUO) dataset~\cite{liu2021dataset} is a combination of six earlier datasets, and contains real seafloor images of scallops, sea urchins (\textit{Echinus}), sea cucumbers (\textit{Holothurian}) and starfish~\cite{liu2021dataset}, with an example image seen in~\cref{fig:DUO_class-specific_performance}.  Prior underwater object detection (UOD) works~\cite{zhao2023yolov7, xu2023systematic, dai2024gated, yuan2023multi, feng2024ceh, zhao2024feb, li2025multi} use this dataset as a benchmark, and demonstrate a clear trend in the relative performance of the different classes~(\cref{tab:DUO_results_previous_studies}).  In these works, the authors attribute the significant under-performance of the scallop class in particular to the scarcity of instances in the training data~\cite{liu2021dataset, zhao2023yolov7, feng2024ceh} or the performance difference is not addressed at all. 
A similar class-specific performance pattern can be observed in the Rethinking general Underwater Object Detection (RUOD) dataset~\cite{fu2023rethinking}, which was carefully collected from various photo-sharing websites to provide data with challenging environmental conditions and a diverse range of ten marine species, including the common categories also contained in DUO. We focus on these four categories and examine some prior studies on the RUOD dataset that report per-class metrics~\cite{chen2025underwater,liu2024marineyolo, yang2024fishdet}, finding that these works also show a performance gap between scallops and the other classes, although this gap is not as pronounced as in DUO. As for DUO, these works do not provide sufficient analysis nor justification for the scallop under-performance.

\begin{table}[t]
  \centering
  \resizebox{\linewidth}{!}{
  \setlength{\tabcolsep}{3pt} 
    \begin{tabular}{@{}lccccc>{\centering\arraybackslash}p{4.4cm}@{}}
      \toprule
      \textbf{Author}  & \textbf{${\text{AP}_{ho}}$} & \textbf{${\text{AP}_{ec}}$} & \textbf{${\text{AP}_{sc}}$} & \textbf{${\text{AP}_{st}}$} & \textbf{mAP} & \textbf{Comment} \\
      \midrule
      \multicolumn{7}{@{}l}{\textit{IoU threshold 0.5:0.95}} \\
      Liu \etal \cite{liu2021dataset} & 
        \colorcell{58.6}{58.6} & \colorcell{69.1}{58.6} & \colorcell{41.3}{58.6} & \colorcell{65.3}{58.6} & 58.6 & ``smallest number of instances" \\
      Xu \etal \cite{xu2023systematic} & 
        \colorcell{59.5}{58.6} & \colorcell{68.5}{58.6} & \colorcell{41.6}{58.6} & \colorcell{64.9}{58.6} & 58.6 & -- \\
      Dai \etal \cite{dai2024gated} &
        \colorcell{68.2}{69.1} & \colorcell{75.2}{69.1} & \colorcell{56.3}{69.1} & \colorcell{76.7}{69.1} & 69.1 & -- \\
      \addlinespace
      \multicolumn{7}{@{}l}{\textit{IoU threshold 0.5}} \\
      Yuan \etal \cite{yuan2023multi} & 
        \colorcell{81.0}{77.0} & \colorcell{84.1}{77.0} & \colorcell{57.6}{77.0} & \colorcell{85.2}{77.0} & 77.0 & -- \\
      Zhao \etal \cite{zhao2023yolov7} & 
        \colorcell{85.3}{84.6} & \colorcell{93.9}{84.6} & \colorcell{64.0}{84.6} & \colorcell{95.2}{84.6} & 84.6 & ``scarcity of scallop instances" \\
      Feng \etal \cite{feng2024ceh} & 
        \colorcell{91.1}{88.4} & \colorcell{93.3}{88.4} & \colorcell{73.3}{88.4} & \colorcell{95.3}{88.4} & 88.4 & ``scarcity of scallop targets" \\
      Zhao \etal \cite{zhao2024feb} & 
        \colorcell{79.6}{82.9} & \colorcell{93.0}{82.9} & \colorcell{65.8}{82.9} & \colorcell{93.1}{82.9} & 82.9 & -- \\
      Li \etal \cite{li2025multi} & 
        \colorcell{84.7}{85.4} & \colorcell{92.6}{85.4} & \colorcell{70.8}{85.4} & \colorcell{93.3}{85.4} & 85.4 & -- \\
      \bottomrule
    \end{tabular}
  }
  \caption{Results reported in previous studies on the DUO dataset.  We show the per-class APs (refer to \cref{sec:exp-frame}), for \textit{ho}: holothurian, \textit{ec}: echinus, \textit{sc}: scallop, \textit{st}: starfish, as well as the mAP. Cell color indicates AP values more than 5\% above the respective mAP in green, AP values within ±5\% in yellow, and AP values more than 5\% below in red. There is persistent under-performance for the scallop class across all studies. We summarize the authors' justification for the class discrepancies, noting that the scallop under-performance is often addressed purely qualitatively or omitted entirely.}
  \label{tab:DUO_results_previous_studies}
  \vspace*{-0.4cm}
\end{table}
Building on these observations, this work presents the first systematic analysis into the under-performance of specific species in underwater object detection. We do not propose a novel architecture in this work, and instead identify common failure modes and species-specific factors, moving beyond the often-cited explanation of data scarcity, and offer key recommendations to guide future research in this area. We investigate two important factors: the dataset characteristics, specifically the class distribution and data quantity; and the influence of inherent visual characteristics of the target objects. While the former can be changed when collecting the data and choosing the samples, the latter needs to be addressed on the algorithmic level. We also decompose the object detection architecture  into the core tasks of localization and classification, to examine which stages are more impacted by these different data challenges. Understanding the underlying reasons for the class-wise detection performance is crucial to design improved object detectors for underwater ecosystems. 

In this work, we present the following contributions:
\begin{enumerate}[itemsep=0pt, parsep=0pt, topsep=0pt]
    \item We systematically decompose the object detection process into localization and classification stages to investigate the causes of class-specific performance disparities, finding that intrinsic visual features play a key role.
    \item We conduct a dedicated localization study, using the TIDE toolkit~\cite{bolya2020tide} to analyze detection errors across different species, and identify that foreground-background discrimination is the main bottleneck, particularly for visually camouflaged species.
    \item We study the classification stage of the object detector, revealing persistent performance gaps under balanced conditions.  Further, we identify inter-class dependencies and a distribution-dependent tradeoff between precision and recall.
    \item We highlight the importance of class-aware performance evaluation and provide recommendations to improve underwater object detector performance, considering common use cases and operational requirements. %
\end{enumerate}

\noindent We will make our evaluation and data processing code available upon acceptance to foster further research in this area.%

\section{Related Works}
\label{sec:rel_works}
Object detection combines the basic tasks of \textit{localization}, \ie defining an object's position using a bounding box, and \textit{classification}, \ie categorizing an object into a specified class, to find and recognize multiple objects within an image~\cite{neha2025classical}. Here, we focus on Underwater Object Detection, and seek to establish the key factors which influence species-specific detection performance. To this end, we first review the unique characteristics and challenges of the underwater domain in \cref{subsec:rel-works-chall}, before summarizing recent underwater object detection approaches in \cref{subsec:rel-works-UOD}, and describing methods and metrics for error evaluation in \cref{subsec:rel-works-fail}.

\subsection{Challenges of the Underwater Domain}
\label{subsec:rel-works-chall}
Transferring generic object detection to marine environments poses unique challenges caused by the characteristics of underwater imagery; these include poor visibility, turbidity, blur, low contrast, light scattering, and color distortion, and complicate feature extraction and foreground-background distinction~\cite{fayaz2022underwater, raine2024reducing, yuan2023multi, er2023research}. These challenges are compounded by the small size of many marine targets, frequent clustering of targets, lack of visual features, occlusion, and unclear object boundaries~\cite{er2023research, feng2024ceh, li2025multi}. Few researchers consider the ability of marine species to camouflage~\cite{er2023research, fu2023rethinking}.

Learning the distinct features of marine species is therefore complicated by in-class variation in appearance and scale, and differences in the angle, distance and speed of the camera~\cite{yuan2023multi, raine2024reducing, chen2024underwater}. While abundant underwater images can be obtained, the annotation effort is very time and cost-intensive and must be performed by domain experts to reduce the risk of noisy and incorrect annotations~\cite{raine2024reducing, chen2024underwater}. Finally, class imbalance is a common issue in underwater data, and is caused by differences in the economic value and rarity of certain species~\cite{liu2021new,xu2023systematic}.

This paper aims to separate the impacts of dataset characteristics \ie quantity and class distribution, from the effects of species-specific visual features and behavioral traits.

\subsection{Underwater Object Detection}
\label{subsec:rel-works-UOD}
Object detectors are typically transferred to the underwater domain as pre-trained detectors fine-tuned on marine datasets~\cite{chen2024underwater}. Popular approaches build upon Fast- and Faster-RCNN~\cite{Girshick_2015_fastRCNN, ren2016fasterRCNN} and YOLO~\cite{Redmon_2016_YOLO}. Recent works have tailored detectors to the unique characteristics of the marine environment~\cite{chen2024underwater}. These include modified architectures with structural changes to the backbone for better feature extraction~\cite{yuan2023multi, cheng2024attention, zhao2024feb, liu2023dsw, feng2024ceh}, improved neck structures for feature fusion, in particular focusing on small targets~\cite{zhao2024feb} and multiple scales~\cite{li2025multi}, and adjusted loss functions, such as IoU-based~\cite{liu2023dsw, feng2024ceh, cheng2024attention} or intentional sample selection~\cite{chen2022swipenet, song2023boosting}. Similarly, different training strategies were proposed to better understand complex underwater environments, prioritizing easy data in early training~\cite{chen2022swipenet} or conversely, focusing on the most difficult samples with the highest loss values first~\cite{wang2024mdi}.

To address the image quality issue, underwater image enhancement is a popular research direction~\cite{wang2023UIE}. However, several studies have already shown that it does not strictly correlate to better performance~\cite{dai2024gated, xu2023systematic, zhao2023yolov7}. Other common data pre-processing includes data augmentation to diversify the training samples for better generalization~\cite{crasto2024class}. Models can be specifically prepared for the underwater domain by simulating overlap, occlusion and blurring in training samples using RoIMix~\cite{lin2020roimix} or by changing position, size and quantity of objects to add minority class examples~\cite{liu2021new}. This directly addresses the common class imbalance problem in underwater data by synthetic oversampling.

While different methods allow majority classes to be under-sampled to create balance \ie by randomly removing majority class samples in RandomUnderSampling (\textit{RUS})~\cite{jin2025weight}, minority classes can be over-sampled by reusing or generating more examples. However, this increases the likelihood that duplicated samples provide insufficient information and cause overfitting~\cite{chen2024underwater}. Additionally, majority and minority class objects are often captured in the same image, so  attempting to over-sample the minority class can also inadvertently increase majority class examples~\cite{crasto2024class}.
Alternatives to manual over/under-sampling include weighted sampling, in which underrepresented classes are assigned higher weights and importance during training~\cite{jin2025weight}.

\subsection{Failure Analysis}
\label{subsec:rel-works-fail}
The object detection performance on single classes is typically measured in Average Precision (AP), which is the area under the precision-recall curve (see \cref{sec:exp-frame}), whereas the overall detector performance calculated by averaging the class APs to obtain the mean AP (mAP).

However, if only the aggregated metrics are used, the localization accuracy %
and potential performance discrepancies across classes remain hidden. The respective influence of False Negative (FN) and False Positive (FP) errors are ignored.  Hence different precision-recall curves can yield the exact same AP score, despite fundamentally different error patterns~\cite{oksuz2018localization}. A detector might have low recall but high precision, making it suitable for risky tasks like autonomous underwater navigation. Conversely, another detector with the same mAP might instead provide high recall but low precision, which is preferred for ecological monitoring of invasive and harmful species, when it is most critical to avoid missing occurrences and to enable early warnings. 

Although these shortcomings are known and some researchers have developed alternative metrics~\cite{oksuz2018localization, ding2018mean, otani2022optimal}, mAP is still the most commonly used metric, which is why we also report the mAP in this work.

\begin{figure*}
    \centering
    \includegraphics[width=0.6\linewidth, clip, trim=0cm 0cm 0cm 0.45cm]{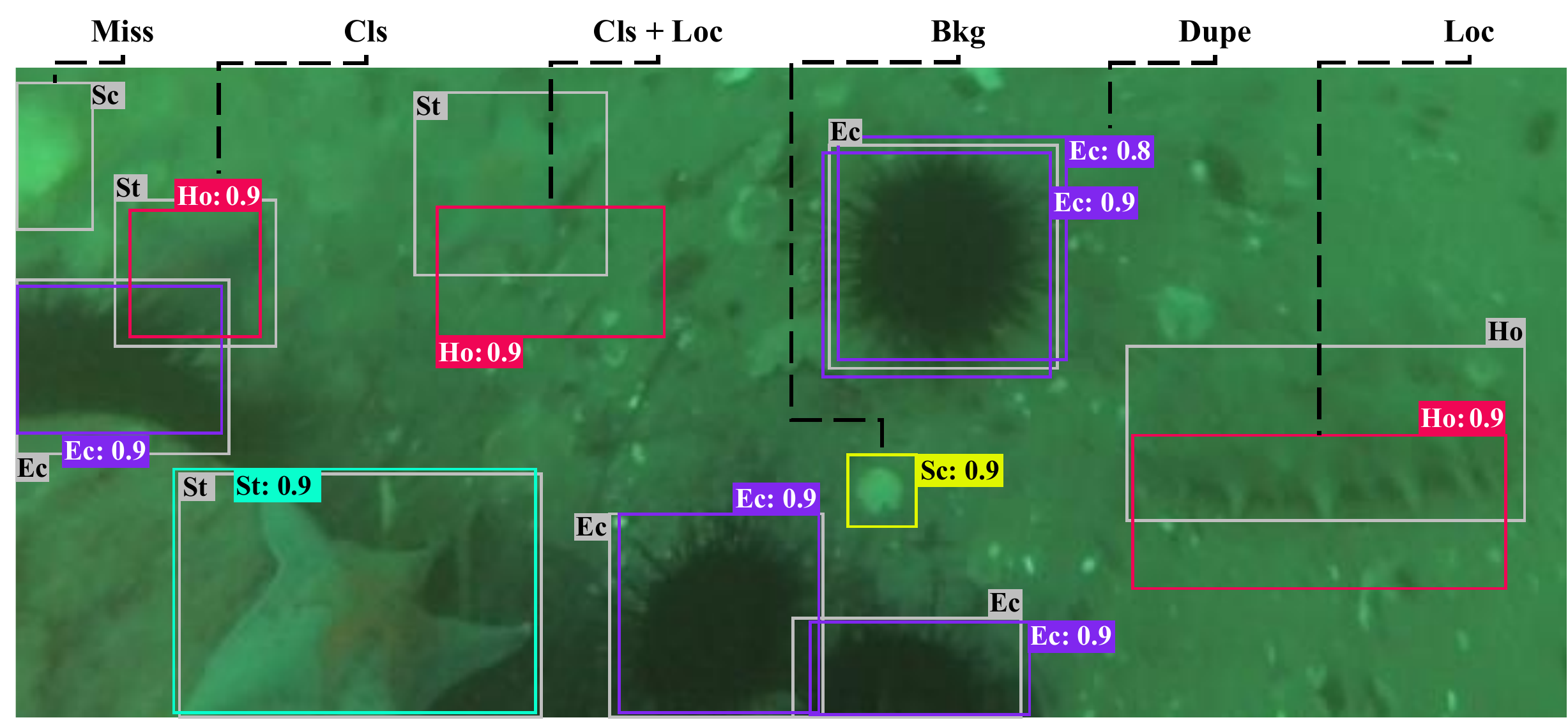}
    \vspace*{-0.1cm}
    \caption{Visualization of TIDE error types on an example DUO image. Ground truth is indicated in grey, predicted bounding boxes in color. The error types are: ``Missed GT'' (completely undetected ground truth, causes FN), ``Classification - Cls'' (correct box location but wrong class, causes FP \& FN), ``Classification and Localization'' (wrong class and insufficient box alignment, causes FP \& FN), ``Background - Bkg'' (background predicted as object, causes FP), ``Duplicate - Dupe'' (duplicate prediction for already matched ground truth, causes FP), and ``Localization - Loc'' (correct class but insufficient IoU overlap with ground truth, causes FP \& FN). Classification, localization, and combined errors are formally counted as false positives in TIDE but also represent false negatives since they leave ground truths unmatched.}
    \label{fig:TIDE_visualization}
    \vspace*{-0.3cm}
\end{figure*}

Nonetheless, additional tools can be applied to gain further insights into the detector performance.  Common tools include  Diagnosis~\cite{hoiem2012diagnosing}, TIDE~\cite{bolya2020tide} and False Negative Mechanisms~\cite{miller2022s}. While Diagnosis categorizes FPs and the FN Mechanisms focus on component failures causing FNs, TIDE proposes six main error types that can either result in FPs, in FNs or both, as explained and visualized in~\cref{fig:TIDE_visualization}. To quantify the impact of each error type on overall performance, TIDE uses delta AP (dAP) by default, which measures the drop in AP caused by each error type in isolation. We utilize TIDE in our localization experiments to evaluate and report the common error types for each class.

Despite the availability of failure analysis tools and known challenges of underwater object detection,  existing studies primarily attribute class-specific performance disparities to data quantity alone~\cite{feng2024ceh, liu2021dataset, zhao2023yolov7}, without systematic investigation of the underlying mechanisms. To our knowledge, no prior work has decomposed underwater object detection performance to isolate the relative contributions of classification versus localization errors across different marine species.  In this work, we perform a systematic investigation of the key object detection stages, and perform failure analysis to isolate the key contributing factors of class-specific performance. This enables us to propose a comprehensive set of recommendations for underwater object detection in different operational scenarios. 

\section{Experimental Framework}
\label{sec:exp-frame}
This section details the general baseline for all our experiments bysummarizing our approach, introducing the dataset of interest and the performance metrics used. 

\textbf{Approach:}
Object detection can be broadly considered as three key stages: (i) distinguishing between the background of an image and objects of interest; (ii) detected objects of interest are tightly localized with a bounding box; and (iii) localized objects of interest are classified. We investigate each of these in turn to identify how class-wise performance is affected by inherent visual features versus characteristics of the dataset. Steps (i) and (ii) are part of our localization analysis (\cref{sec:localization}), whereas (iii) is examined in a classification-focused study (\cref{sec:classification}). In every step we consider the influence of data quantity and distribution versus inherent class-specific properties.

\textbf{Base Dataset:}
We perform our experiments with a primary focus on the DUO dataset \cite{liu2021dataset}. DUO provides a standardized benchmark and assembles images from the Underwater Robot Professional Contest from 2017 to 2020, as well as the Underwater open-sea farm object Detection Dataset (UDD)~\cite{liu2021new}. In its original form, DUO provides 6,671 training and 1,111 test images, with 50,156 labeled echinus (67\%), 7,887 holothurian (11\%), 1,924 scallop (3\%) and 14,548 starfish objects (19\%). We generate several subsets from the original DUO for our localization and classification studies, as detailed in \cref{subsec:local-datasets,subsec:cls-datasets}. To confirm the generalizability of the results, we also apply our experimental framework to a modified version of the RUOD dataset~\cite{fu2023rethinking}, called RUOD-4C, which only includes the images that contain echinus, holothurian, scallop and/or starfish annotations, \ie 11,292, 7,626, 7,411 and 8,118 objects. The RUOD-4C class distribution is relatively even (33\%, 22\%, 22\%, 23\% respectively). While the original RUOD consists of 9,800 training and 4,200 testing images, these numbers are reduced to 3,013 and 1,255 images in our RUOD-4C.

\textbf{Evaluation Metrics:}
We use the common object detection metric mAP for the localization experiments, and precision and recall for classification. The mAP is calculated by taking the average precision, $\text{AP} = \int_0^1 p(r) \, dr$, and averaging over the classes $C$:
$\text{mAP} = \frac{1}{C} \sum_{i=1}^{C} \text{AP}_i$.

Additionally, we specifically evaluate bounding box accuracy per class by computing the center and area error.
The center offset  $O = \left\| \mathbf{c}_p - \mathbf{c}_g \right\|_2$ measures the Euclidean distance between the centers of the predicted ($\mathbf{c}_p$) and the ground truth bounding box ($\mathbf{c}_g$). %
The area error $E$ measures the relative deviation between predicted and ground truth bounding box areas as a percentage.

To evaluate classification performance, we look at the proportion of correct predictions (true positives, $TP$) out of all predictions made, which is given as precision $p = \frac{TP}{(TP + FP)}$, and the proportion of correctly predicted objects out of all true objects, called recall or True Positive Rate (TPR), $\text{TPR} = \frac{TP}{(TP + FN)}$. We also consider the False Negative Rate (FNR), as the proportion of missed targets out of the total, $\text{FNR} = \frac{FN}{(TP + FN)}$, and the False Discovery Rate, which is the ratio of false positives to total predictions, $\text{FDR} = \frac{FP}{(FP + TP)}$.  Finally, the F1 Score balances precision and recall: $F1 = 2 \cdot \frac{\text{Precision} \cdot \text{Recall}}{\text{Precision} + \text{Recall}}$.

\section{Localization Study}
\label{sec:localization}

This section describes our localization analysis by providing an overview of the methodology (\cref{subsec:local-method}), datasets (\cref{subsec:local-datasets}) and implementation details (\cref{subsec:local-implement}). We then discuss our results evaluating how well each class is distinguished from the background (\cref{subsec:res-foreVSback}) and our results evaluating  bounding box localization (\cref{subsec:res-bbox}).

\subsection{Method}
\label{subsec:local-method}
In our localization experiments, we eliminate the classification stage interference by creating single-class versions of the datasets and train object detectors on them to analyze class-specific localization challenges independently. The focus lies on assessing foreground-background confusion and examining bounding box quality. We also leverage TIDE~\cite{bolya2020tide} to gain detailed insights into the underlying error types. 

We evaluate both the effect of data quantity and the impact of including or excluding unannotated images. We carefully consider possible operational requirements of underwater monitoring settings, and use these to inform the recommendations made.

\subsection{Datasets}
\label{subsec:local-datasets}
  We obtain \textit{single-class} DUO/RUOD-4C versions using the original images but modifying the annotations to omit the labels for all but one species at a time. A random 20\% from the training split is selected for the validation split. To obtain further subsets, we select all ``empty" images first and then add images containing the target species until reaching the desired instance count. Since images may contain multiple objects of the same class, we include all instances of the target species when an image is selected. The subsets include:
    (i) balanced sets in which we use the minority class (scallop) count as the target, so that each single-class dataset contains an equal number of training objects;
    (ii) reduced sets with the instance count decreased to 75\%, 50\% and 25\% of the balanced amount in each single-class dataset; and
    (iii) a special single-class DUO dataset (\textit{limited scallop}) which does not include any ``empty" images but consists only of images that actually contain the target of interest, scallop, for comparison purposes. This is specific to DUO as the class-imbalance is more pronounced and hence, the difference between the sets is greater.

\subsection{Implementation}
\label{subsec:local-implement}

\begin{figure}[t]
  \centering

  \includegraphics[width=0.95\linewidth, trim=100 3 100 20, clip]{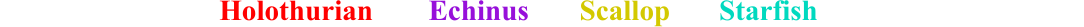}

  \begin{subfigure}[t]{0.49\linewidth}
    \centering
    \includegraphics[width=0.95\linewidth]{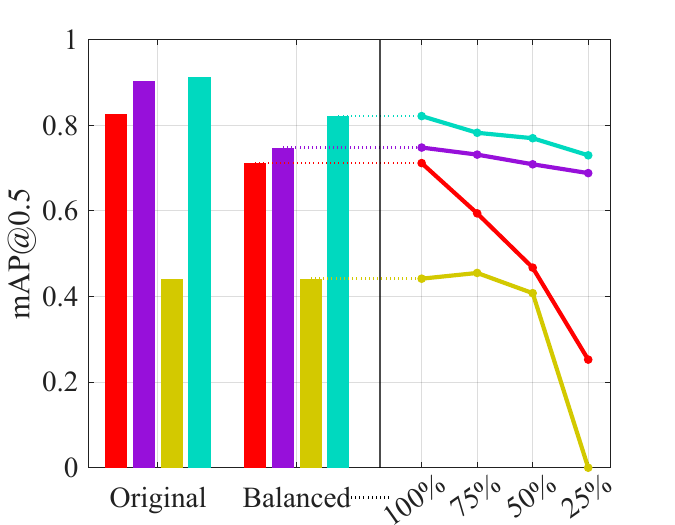}
    \caption{DUO}
    \label{fig:duo_map}
  \end{subfigure}%
  \hfill
  \begin{subfigure}[t]{0.49\linewidth}
    \centering
    \includegraphics[width=0.96\linewidth]{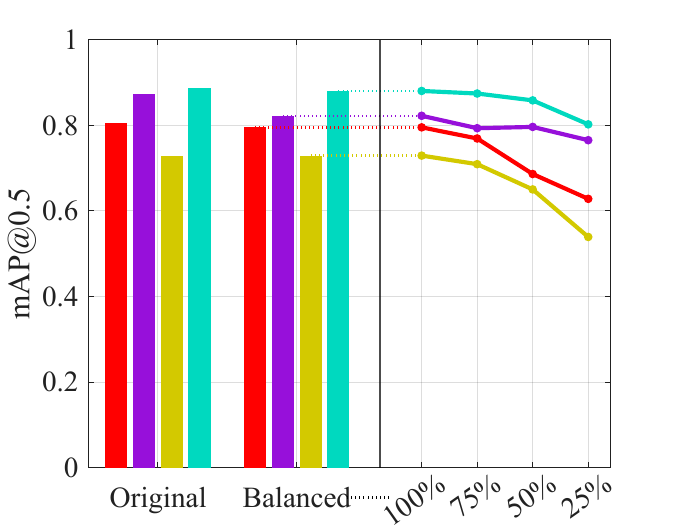}
    \caption{RUOD-4C}
    \label{fig:ruod-4c_map}
  \end{subfigure}
  \vspace*{-0.2cm}
  \caption{Per class localization performance compared across datasets. Bar charts indicate mAP@0.5 for the originally imbalanced and balanced data, line charts represent gradually reduced sets. For both DUO (a) and RUOD-4C (b), scallop performance is the same in the original and balanced sets (all classes in the balanced set are downsampled to the scallop instance count), but still lower than all other classes, despite the same number of instances. Further reductions of the balanced set affect classes differently: starfish and echinus are impacted less by the decrease, whereas holothurian and scallop exhibit significant reduction in mAP.}
  \label{fig:loc_mAP_full_balanced_reduced}
  \vspace*{-0.4cm}
\end{figure}

For the localization experiments, we used Ultralytics YOLO11n\footnote{\url{https://github.com/ultralytics/ultralytics}} as the object detector, pretrained on COCO. The best model is saved based on the validation performance and evaluated on the test set with confidence threshold 0.25 and IoU set to 0.7 (refer to Supp.~Mat.~for additional implementation details). To verify that our findings are not specific to YOLO11n, we repeat the localization study with an SSD detector and obtained similar results, as detailed in the Supp.~Mat.

\subsection{Results: Foreground-Background Separation}
\label{subsec:res-foreVSback}
Our localization experiments reveal that class performance disparities persist even when the class distribution is balanced.  As the left side of~\cref{fig:duo_map} illustrates, the single-class DUO versions show the drastic performance gaps that have motivated our work -- mAP@0.5 scores for echinus and starfish greater than 90\%, followed by holothurian at 82.5\%, but substantially lower scallop performance at  44\% -- yet, even when all datasets are balanced to contain the same number of training labels, these performance gaps remain evident. The difference between scallop and holothurian for mAP@0.5 (the smallest gap) remains 27\%, clearly demonstrating that the model struggles with scallop localization regardless of the quantity of training samples. This is the most major performance difference of all three detector tasks, and hints at challenges due to the inherent visual features.

The right half of~\cref{fig:duo_map} shows that when data quantity is gradually decreased in our balanced setup, the holothurian and scallop classes are more affected by the number of sample points in the training data and sensitive to reductions. While the downwards trend generally highlights the benefits of larger training datasets, the varying rates of performance change again hint at inherent data characteristic differences.

\cref{fig:ruod-4c_map} confirms the same results on RUOD-4C. Although the scallop under-performance is less pronounced with a 7\% gap between balanced holothurian and scallop mAPs, all insights gained from DUO remain valid.

\begin{table}
\centering
\setlength{\tabcolsep}{1pt}
\renewcommand{\arraystretch}{1.4}
\scriptsize
\begin{tabular}{>{\raggedright\arraybackslash}p{0.5cm} 
                >{\centering\arraybackslash}p{1.215cm}
                >{\centering\arraybackslash}p{1.215cm}
                >{\centering\arraybackslash}p{1.215cm}
                >{\centering\arraybackslash}p{1.215cm}
                >{\centering\arraybackslash}p{1.215cm}
                >{\centering\arraybackslash}p{1.215cm}}
\toprule
\ & \multicolumn{3}{c}{\textbf{Original Distribution (Imbalanced)}} & \multicolumn{3}{c}{\textbf{Balanced Sets}} \\
\cmidrule(lr){2-4} \cmidrule(lr){5-7}
 & TPR & FDR & FNR & TPR & FDR & FNR \\
\midrule
Ho & 
\colormap{77}{76.7}\,\textbf{/}\,\colormap{79}{78.7} & \colormap{22}{21.9}\,\textbf{/}\,\colormap{28}{27.8} & \colormap{23}{23.3}\,\textbf{/}\,\colormap{21}{21.3} & 
\colormap{57}{56.8}\,\textbf{/}\,\colormap{76}{76.4} & \colormap{19}{19.2}\,\textbf{/}\,\colormap{25}{25.1} & \colormap{43}{43.2}\,\textbf{/}\,\colormap{24}{23.6} \\

Ec & 
\colormap{88}{88.0}\,\textbf{/}\,\colormap{90}{89.5} & \colormap{14}{14.2}\,\textbf{/}\,\colormap{30}{30.1} & \colormap{12}{12.0}\,\textbf{/}\,\colormap{11}{10.5} & 
\colormap{52}{51.8}\,\textbf{/}\,\colormap{83}{82.6} & \colormap{4}{4.0}\,\textbf{/}\,\colormap{31}{30.8} & \colormap{48}{48.2}\,\textbf{/}\,\colormap{17}{17.4} \\

St & 
\colormap{88}{88.0}\,\textbf{/}\,\colormap{88}{87.8} & \colormap{17}{17.0}\,\textbf{/}\,\colormap{23}{23.1} & \colormap{12}{12.0}\,\textbf{/}\,\colormap{12}{12.2} & 
\colormap{68}{68.3}\,\textbf{/}\,\colormap{86}{85.6} & \colormap{8}{8.4}\,\textbf{/}\,\colormap{19}{19.1} & \colormap{32}{31.7}\,\textbf{/}\,\colormap{14}{14.4} \\

Sc & 
\colormap{24}{24.0}\,\textbf{/}\,\colormap{67}{67.1} & \colormap{36}{35.8}\,\textbf{/}\,\colormap{31}{30.7} & \colormap{76}{76.0}\,\textbf{/}\,\colormap{33}{32.9} & 
\colormap{24}{24.0}\,\textbf{/}\,\colormap{67}{67.1} & \colormap{36}{35.8}\,\textbf{/}\,\colormap{31}{30.7} & \colormap{76}{76.0}\,\textbf{/}\,\colormap{33}{32.9} \\

Sc\textsubscript{\textit{lim}} & 
\colormap{74}{74.2}\,\textbf{/}\,----- & \colormap{76}{75.6}\,\textbf{/}\,-----  & \colormap{26}{25.8}\,\textbf{/}\,-----  & 
 &  &  \\

\bottomrule
\end{tabular}
 \caption{DUO\,/\,RUOD-4C True Positive Rate (TPR), False Discovery Rate (FDR), and False Negative Rate (FNR) per class under two training setups (refer to Sec.~\ref{sec:exp-frame} for metric definitions). 
 Most scallop instances are undetected (76\% FNR). Despite increases in FN for Holothurian, Echinus and Starfish on the balanced set, their recall remains above the scallop value (24\%). Training without background-only images (Scallop\textsubscript{\textit{limited}} row) increases the recall by 50.2\%, but at the expense of an increase of 39.8\% in the FDR.}
 \label{tab:loc_TP_FP_FN}
\vspace*{-0.5cm}
\end{table}
\begin{figure*}
    \centering
    \begin{subfigure}{0.22\textwidth}
        \includegraphics[width=\linewidth]{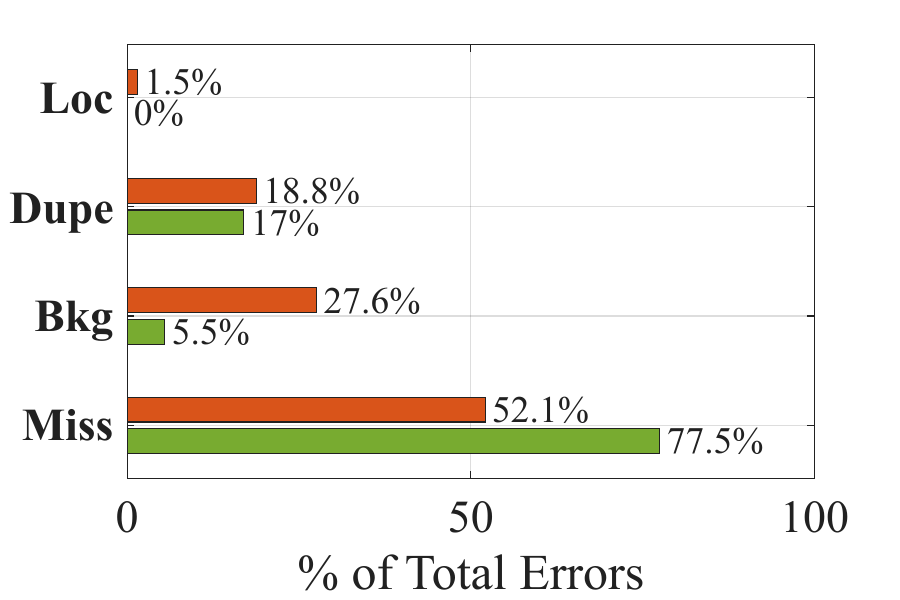}
        \caption{\scriptsize Holothurian}
    \end{subfigure}
    \hfill
    \begin{subfigure}{0.22\textwidth}
        \includegraphics[width=\linewidth]{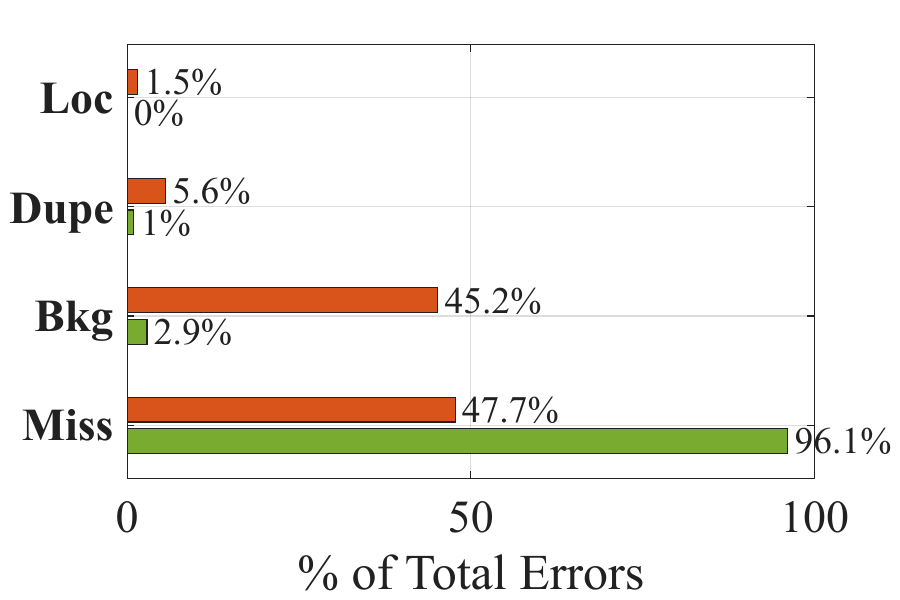}
        \caption{\scriptsize Echinus}
    \end{subfigure}
    \hfill
    \begin{subfigure}{0.22\textwidth}
        \includegraphics[width=\linewidth]{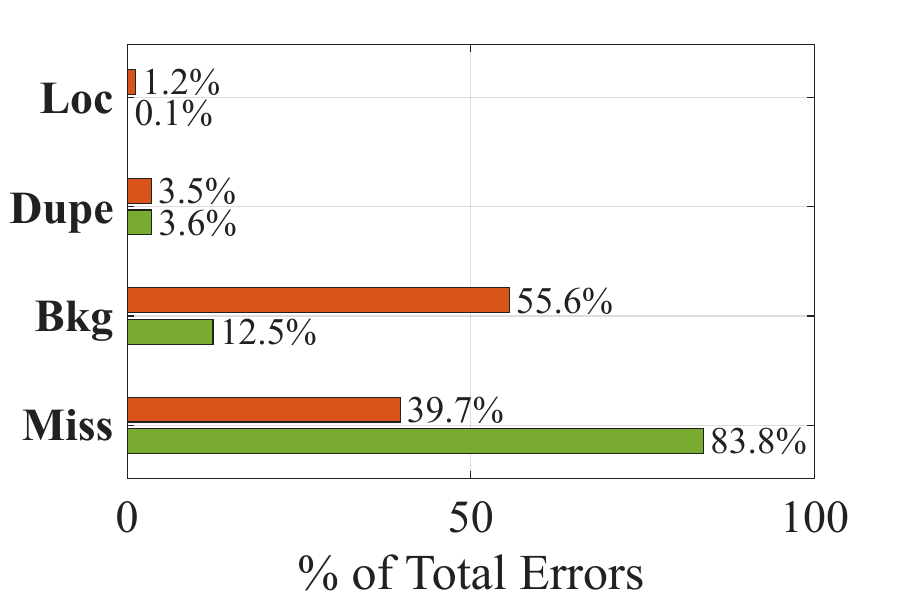}
        \caption{\scriptsize Starfish}
    \end{subfigure}
    \hfill
    \begin{subfigure}{0.22\textwidth}
        \includegraphics[width=\linewidth]{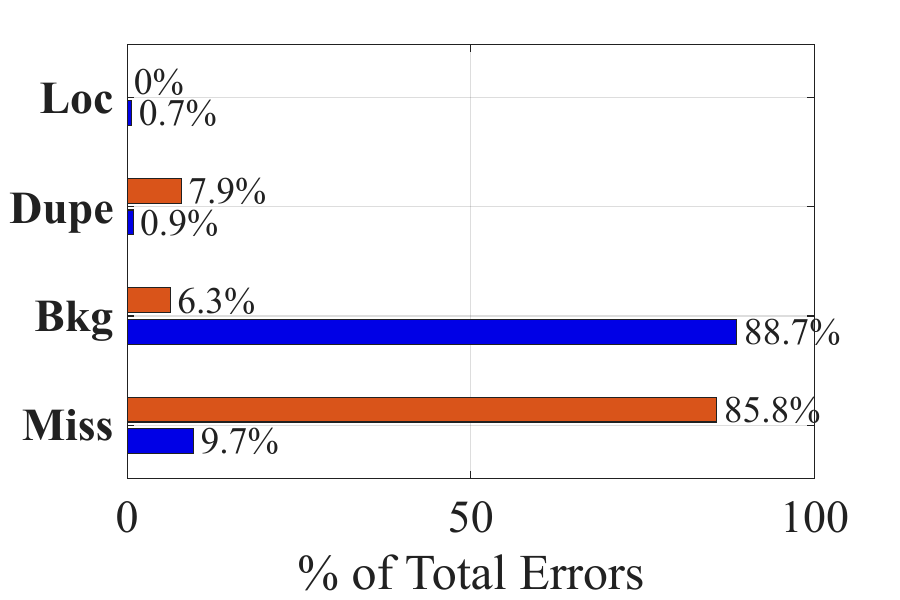}
        \caption{\scriptsize Scallop}
    \end{subfigure}

    \caption{Distribution of TIDE error types across all classes for the original (red) and balanced (green) DUO datasets. The imbalanced versions exhibit higher rates of FPs, and specifically background error, for Starfish, echinus and holothurian classes.  In contrast, balanced datasets make the error profiles for these classes more similar to the scallop class, \ie there are more missed ground truth detections. Only training on object-containing images, as seen in the Scallop Limited bars (blue), reduces misses but massively increases background error.}
    \label{fig:tide_error_breakdown_type}
    \vspace*{-0.4cm}
\end{figure*}

\cref{tab:loc_TP_FP_FN} breaks down the true positive (recall), false discovery and false negative rates under imbalanced and balanced conditions. The results reveal that scallops generally suffer from the worst recall and accordingly are most affected by FNs. In DUO, recall is extremely poor  with only 24\% of true scallops being detected, compared to other classes with recall above 75\% for imbalanced data. The balanced sets resulted in decreased recall by 19.9\%, 36.2\% and 19.7\% for holothurian, echinus and starfish, respectively, but even that performance decline is not enough to close the gap. Despite the equal number of training instances, the scallop FN rate (76\%) is much higher compared to the other classes, where only 31.7-48.2\% are being missed \ie scallops are most often mistaken as part of the background. In the imbalanced DUO sets, the FDR is also highest for the scallop class at 35.8\%, as compared to 14-21\% for the other classes.  When the sets are balanced, the FDR reduces by 2.7\% to 19.2\% for holothurian, by 10.1\% to 4.0\% for echinus, and by 8.5\% to 8.4\% for starfish, increasing the gap to the high scallop FDR to a maximum difference of 16.6\%.

In RUOD-4C, the performance disparities are not as strong but are still evident. The FDRs are generally higher in RUOD-4C than in DUO, with scallops and holothurians being prone to false positive detections most. The scallop class notably shows the lowest recall and highest FNR in both class distributions.

Qualitative inspection confirms the high background confusion, with scallops being the hardest to spot due to their ability to camouflage. In~\cref{fig:TIDE_visualization}, the \textit{missed} example is a small scallop blending into the background, resembling a faint spot, easily mistaken for scattered light or a rock. The \textit{background} error shows a rock falsely detected as a scallop, highlighting their visual similarity. In contrast, other classes are more distinct -- echinus are dark and spiny, starfish have clear five-arm symmetry, and holothurians are elongated with surface spikes. Thus, scallops are arguably the most difficult class to distinguish from background clutter.

Overall, background confusion~\ie~misinterpreting background as an object or an object as background, is the biggest issue. This is evident~\cref{fig:tide_error_breakdown_type}, highlighing background errors (Bkg) that cause FPs and missed GT errors (Miss) causing FNs as the most prominent error types that occur across all classes. With the original DUO instance count per class, starfish, echinus and holothurian still result in a considerable amount of Bkg's, while the balanced versions yield high rates of missed GT errors. Of all the classes, scallops have the highest error rate relative to the number of test objects.

An important consideration when creating our datasets was whether to include background-only images. The comparisons between full scallop datasets and limited scallop-only datasets of DUO in~\cref{tab:loc_TP_FP_FN} and~\cref{fig:tide_error_breakdown_type} reveal that including diverse background images (even those without scallops) improves model generalization and heavily reduces false positives, though recall is reduced compared to training only on scallop-containing images.

\subsection{Results: Bounding Box Placement}
\label{subsec:res-bbox}
\begin{figure}[t]
  \centering
  \begin{subfigure}[t]{0.5\columnwidth}
    \centering
    \includegraphics[width=\linewidth]{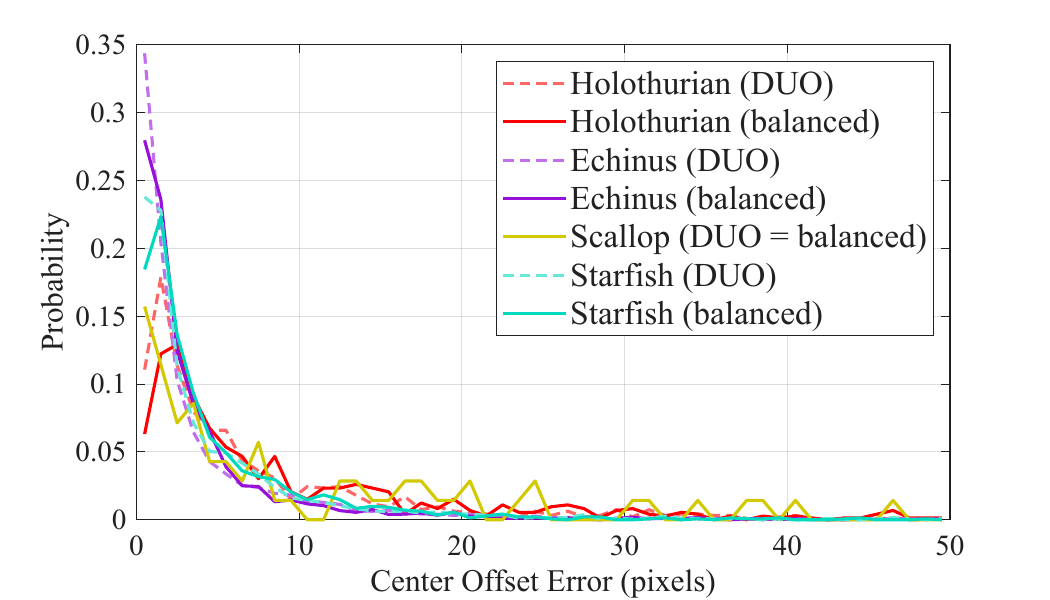}
    \caption{BBox Center Offset}
    \label{fig:map_05}
  \end{subfigure}%
  \hfill
  \begin{subfigure}[t]{0.5\columnwidth}
    \centering
    \includegraphics[width=\linewidth]{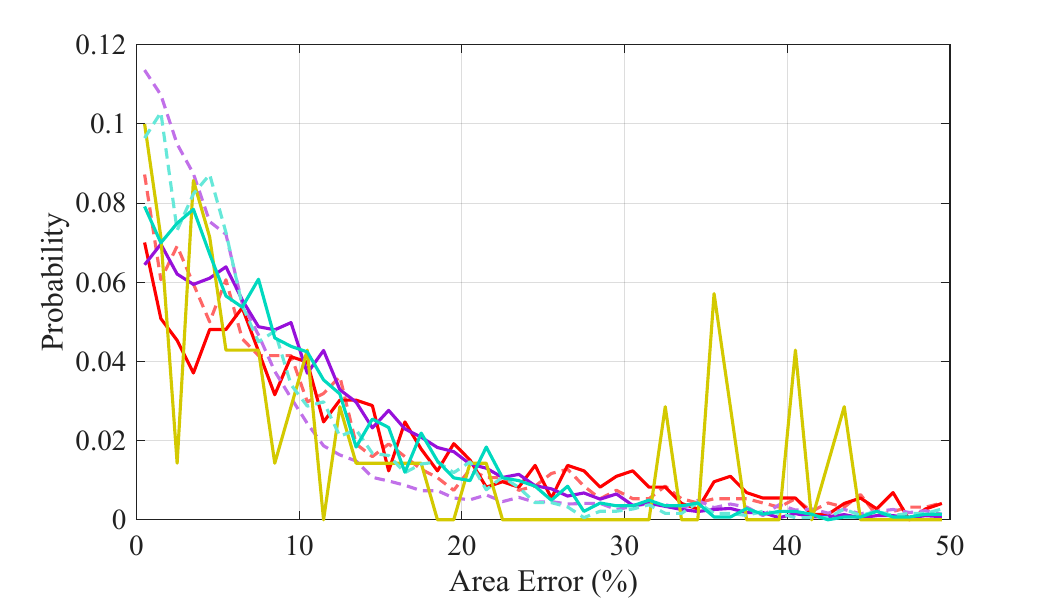}
    \caption{BBox Area Error}
    \label{fig:map_5095}
  \end{subfigure}
  \vspace*{-0.1cm}
  \caption{Accuracy of predicted bounding boxes for DUO.  (a) Bounding box center offset, with minor deviations but overall similar distribution across classes and datasets. (b) The bounding box area error, which highlights larger fluctuations for the small scallop objects but a similar trend overall.}
  \label{fig:bbox_metrics}
  \vspace*{-0.5cm}
\end{figure}

\cref{tab:map_deviation} indicates that once an object is distinguished from the background, further localization performance differences are not significant. 
\begin{table}
\centering
  \setlength{\tabcolsep}{4.5pt}
  \scriptsize
  \begin{tabular}{lcccr}
    \toprule
    & \multicolumn{2}{c}{\textbf{mAP@0.5}} 
    & \multicolumn{2}{c}{\textbf{mAP@0.5:0.95}} \\
    \cmidrule(lr){2-3} \cmidrule(lr){4-5}
    \textbf{Class} & \textbf{Value} & \textbf{Deviation} 
                   & \textbf{Value} & \textbf{Deviation} \\
    \midrule
    Holothurian & $0.71$\,\textbf{/}\,$0.80$ &  $-0.11$\,\textbf{/}\,$-0.08$ & $0.50$\,\textbf{/}\,$0.51$ &  $-0.15$\,\textbf{/}\,$-0.06$ \\
    Echinus     & $0.75$\,\textbf{/}\,$0.82$ & $-0.07$\,\textbf{/}\,$-0.06$ & $0.60$\,\textbf{/}\,$0.49$ & $-0.05$\,\textbf{/}\,$-0.08$ \\
    Starfish    & $0.82$\,\textbf{/}\,$0.88$ &  $0.00$\,\textbf{/}\,$0.00$ & $0.65$\,\textbf{/}\,$0.57$ &  $0.00$\,\textbf{/}\,$0.00$ \\
    Scallop     & $0.44$\,\textbf{/}\,$0.73$ & $-0.38$\,\textbf{/}\,$-0.15$ & $0.35$\,\textbf{/}\,$0.49$ & $-0.30$\,\textbf{/}\,$-0.08$ \\
    \bottomrule
  \end{tabular}
   \caption{DUO\,/\,RUOD-4C per-class mAP results on the balanced dataset and the deviation from the best‐performing class. The scallop class under-performance is less pronounced with higher IoU thresholds.}
   \label{tab:map_deviation}
  \vspace*{-0.5cm}
\end{table}
 Specifically, when requiring better bounding box accuracy via a higher IoU threshold, the performance discrepancy becomes smaller between classes. This implies that localization precision is similar across classes, and the main cause of difference at lower thresholds is due to detection difficulty, not box alignment. This is underlined in~\cref{fig:tide_error_breakdown_type}: the balanced datasets have rarely any localization errors, \ie if the object is not missed, the predicted bounding box usually is accurate enough. While there are some minor differences in the precision of bounding boxes across classes (\cref{fig:bbox_metrics}), with the smaller and visually less distinct scallop objects having slightly larger center offsets and higher area errors more frequently, all classes follow a very similar error distribution regardless of imbalanced or balanced setup.
 \vspace*{-0.1cm}

\section{Classification Study}
\label{sec:classification}

This section describes our analysis on the classification stage of object detection. \cref{subsec:cls-method} describes the methodology,  we explain the datasets in \cref{subsec:cls-datasets}, and implementation details are in \cref{subsec:cls-implement}. \cref{subsec:res-cls}  highlights our findings on  what factors impact the classification of detected targets.

\begin{table*}
\centering
\renewcommand{\arraystretch}{1.15}
\setlength{\tabcolsep}{3pt}
\scriptsize
\begin{tabularx}{\textwidth}{>{\raggedright\arraybackslash}l *{12}{>{\centering\arraybackslash}X}}
\toprule
& \multicolumn{4}{c}{\textbf{Precision}} & \multicolumn{4}{c}{\textbf{Recall}} & \multicolumn{4}{c}{\textbf{F1 Score}} \\
\cmidrule(lr){2-5} \cmidrule(lr){6-9} \cmidrule(lr){10-13}
\textbf{Dataset Version} & $\mathrm{Ho}$ & $\mathrm{Ec}$ & $\mathrm{Sc}$ & $\mathrm{St}$ 
& $\mathrm{Ho}$ & $\mathrm{Ec}$ & $\mathrm{Sc}$ & $\mathrm{St}$ 
& $\mathrm{Ho}$ & $\mathrm{Ec}$ & $\mathrm{Sc}$ & $\mathrm{St}$ \\
\midrule
\textit{Cls Reference Set}      & \textbf{95.2}\,\textbf{/}\,\textbf{95.4} & 99.1\,\textbf{/}\,\textbf{98.1} & 94.3\,\textbf{/}\,\textbf{98.2} & 98.2\,\textbf{/}\,97.8 & 97.0\,\textbf{/}\,96.7 & 99.4\,\textbf{/}\,\textbf{98.3} & 91.6\,\textbf{/}\,97.7 & 96.4\,\textbf{/}\,\textbf{97.0} & 96.1\,\textbf{/}\,\textbf{96.0} & 99.3\,\textbf{/}\,\textbf{98.2} & 92.9\,\textbf{/}\,\textbf{98.0} & 97.3\,\textbf{/}\,\textbf{97.4} \\
\textit{Imbalanced Minimum}    & 88.4\,\textbf{/}\,95.3 & 97.8\,\textbf{/}\,97.9 & 77.6\,\textbf{/}\,97.7 & 95.9\,\textbf{/}\,97.3 & 89.7\,\textbf{/}\,96.1 & 98.8\,\textbf{/}\,98.2 & 70.3\,\textbf{/}\,97.6 & 92.1\,\textbf{/}\,96.4 & 89.0\,\textbf{/}\,95.7 & 98.3\,\textbf{/}\,98.0 & 73.3\,\textbf{/}\,97.7 & 94.0\,\textbf{/}\,96.8 \\
\textit{Balanced Full} & 84.2\,\textbf{/}\,93.7 & 99.0\,\textbf{/}\,97.9 & 66.4\,\textbf{/}\,98.1 & 95.5\,\textbf{/}\,\textbf{97.9} & 95.2\,\textbf{/}\,\textbf{97.1} & 96.9\,\textbf{/}\,97.5 & \textbf{96.2}\,\textbf{/}\,97.6 & 92.9\,\textbf{/}\,96.3 & 89.3\,\textbf{/}\,95.3 & 97.9\,\textbf{/}\,97.8 & 78.6\,\textbf{/}\,97.7 & 94.2\,\textbf{/}\,97.1 \\
Weighted Sampling & 95.0\,\textbf{/}\,95.3 & \textbf{99.3}\,\textbf{/}\,97.9 & \textbf{94.6}\,\textbf{/}\,97.8 & \textbf{98.7}\,\textbf{/}\,\textbf{97.9} & \textbf{97.9}\,\textbf{/}\,96.5 & \textbf{99.5}\,\textbf{/}\,\textbf{98.3} & 93.4\,\textbf{/}\,\textbf{97.8} & \textbf{96.6}\,\textbf{/}\,96.5 & \textbf{96.4}\,\textbf{/}\,95.9 & \textbf{99.4}\,\textbf{/}\,98.1 & \textbf{94.0}\,\textbf{/}\,97.8 & \textbf{97.6}\,\textbf{/}\,97.2 \\
\bottomrule
\end{tabularx}
\caption{DUO\,\textbf{/}\,RUOD-4C classification precision, recall and F1 score (refer to Sec.~\ref{sec:exp-frame} for definitions) across different dataset distributions. Results on DUO highlight a precision-recall tradeoff: balanced datasets improve scallop recall, imbalanced datasets result in higher precision. Weighted sampling mitigates scallop under-performance best. For RUOD-4C, metrics are consistently high across classes and data distributions, reflecting the near-balanced original class distribution.}
\label{tab:cls-results}
\vspace*{-0.5cm}
\end{table*}

\subsection{Method}
\label{subsec:cls-method}
In the classification experiments, we focus purely on how well the query targets are assigned to the correct class. To isolate this stage from localization effects, we extract single-object image crops that simulate the output of a bounding box regressor and feed them directly to a classifier. This setup enables controlled manipulation of the training samples to establish the impact of data quantity and class imbalance.

For each experiment, we fine-tune a pre-trained classifier three times with different random seeds to minimize influence of random variation. Across experiments, the model architecture, hyperparameters, and test set remain consistent. We report results as the average across the three runs.%

\subsection{Datasets}
\label{subsec:cls-datasets}
Given that DUO and RUOD-4C were designed for object detection, we modified the data for our classification experiments to obtain single-object images. First, we cropped square images of every object with at least 10px padding around the bounding box.  We randomly selected 20\% of the training data for validation and removed all cropped images where the original bounding box overlapped by more than 10\% with any other annotated object, thereby avoiding crops that contain multiple species. We call this cleaned subset the \textit{classification reference set}, which maintains a similar class distribution to the original data in DUO (\cref{sec:exp-frame}),~\ie~ there is a pronounced class imbalance. In RUOD-4C, the holothurian class loses proportionally a greater quantity of images, making holothurian the smallest class in this scenario.

The classification reference sets serve as the basis for further subsets: 
    (i) balanced sets with randomly down-sampled classes so that all provide the same data quantity as the minority class, referred to as \textit{Balanced Full};
    (ii) reduced sets in which a) the total data quantity from the classification DUO reference set is decreased for an imbalanced minimum set of the same size as the balanced set maintaining class distribution, referred to as \textit{Imbalanced Minimum} and b) the total data quantity from the balanced set is gradually decreased keeping equal instance count per class; and
    (iii) differently distributed sets where only samples of one class at a time are gradually reduced. 
We also conduct an experiment on the classification reference sets with weighted sampling to address class imbalance.

\subsection{Implementation}
\label{subsec:cls-implement}
We used a ResNet-18~\cite{he2016deep} pre-trained on ImageNet as our classifier implemented through PyTorch.  The same experiments were repeated with MobileNetV2 and VisionTransformer-B/16 as classifiers and confirmed the results obtained with ResNet-18 (for further details, please refer to the Supp.~Mat.).

\subsection{Results}
\label{subsec:res-cls}
\begin{figure}[t]
  \centering
  \includegraphics[width=0.85\linewidth]{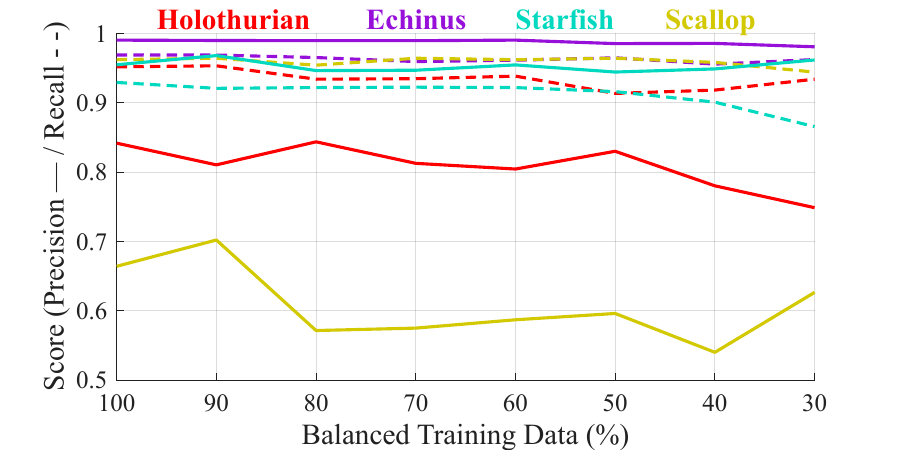}
  \caption{Per class classification performance on incrementally reduced balanced DUO subsets (Solid lines: Precision; dashed lines: Recall). Recall is consistently high across all classes but there are significant gaps in precision, with the scallop class experiencing the largest fluctuations with reduced data.}
  \label{fig:cls_balanced_p_r}
  \vspace*{-0.6cm}
\end{figure}

\cref{tab:cls-results} demonstrates very good classification performance in general. The only major disparities are found in the scallop precision values for DUO. In RUOD-4C, all metrics are generally high and very similar across classes and across data versions. Considering the original RUOD-4C class distribution is nearly balanced (\cref{sec:exp-frame}), it is not surprising that imbalanced vs. balanced, full vs. minimal or unweighted vs. weighted data show no significant differences. However the pronounced class imbalance in DUO requires in-depth analysis: examination of the DUO performance reveals a fundamental precision-recall tradeoff based on the training distribution. For scallops, imbalanced training yields higher precision but lower recall, while balanced training increases recall by 4.6\% (from 91.6\% to 96.2\%) but significantly reduces precision by 27.9\% from 94.3\% to 66.4\%. This pattern is most pronounced for scallops, with other classes showing greater robustness across training strategies. 
The performance gaps in DUO become almost negligible when weighted sampling is applied. When the data is quantitatively balanced (\textit{Balanced Full}), all classes achieve comparable and high recall but significant performance disparities are evident in precision: \cref{fig:cls_balanced_p_r} reveals that even with balanced training, scallops achieve 55-70\% precision, vs. $>$95\% for echinus and starfish, demonstrating intrinsic visual challenges. %

This indicates that while there are benefits for larger training datasets, classification performance differences caused by inherent data characteristics persist. The scallop class exhibits both the lowest overall precision, and the largest performance fluctuations as data is reduced, suggesting that scallop features are inherently less distinctive and more sensitive to data availability.

\begin{figure}[t]
  \centering
  \vspace*{-0.1cm}
  \includegraphics[width=0.84\linewidth, clip, trim={0 0.2cm 0 0.65cm}]{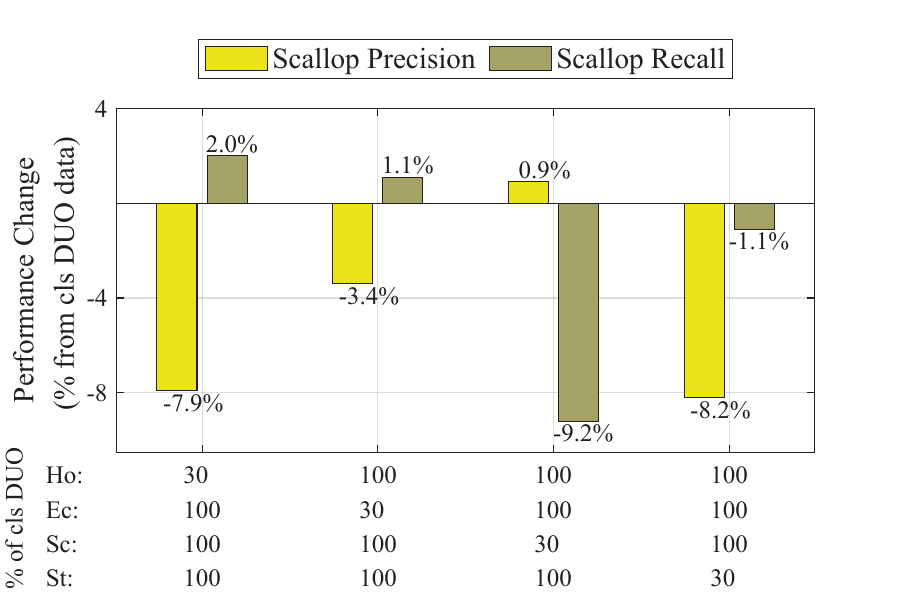}
  \caption{Effect of data reductions on scallop performance in DUO. From the classification reference set (100\%), training samples are reduced to 30\% one class at a time. Scallop precision declines as training data for other classes (particularly starfish and holothurian) is reduced, despite keeping scallop data constant. The model relies on sufficient negative examples to understand class boundaries.}
  \label{fig:cls_reduced_class_data_effect}
\vspace*{-0.6cm}
\end{figure}

\cref{fig:cls_reduced_class_data_effect} shows that reducing training data for other classes in DUO significantly impacts scallop classification performance, even when scallop training data remains constant. Interestingly, reducing starfish data to 30\% causes an 8.2\% drop in scallop precision and a 1.1\% reduction in scallop recall, while reducing holothurian data results in a 7.9\% precision decrease for scallops.

This interdependence suggests that the model learns scallop identity partially through negative examples, \ie understanding what scallops are not. When other classes have insufficient representation, the model becomes less certain about class boundaries, leading to more false positive scallop predictions and reduced precision.

\vspace{-0.1cm}
\section{Discussion and Conclusion}
\label{sec:discussion_conclusion}

This work demonstrates that performance disparities in underwater object detection are not simply artifacts of imbalanced training data, but reflect fundamental differences in the visual characteristics and detectability of marine species. Through systematic decomposition of object detection into localization (foreground versus background followed by bounding box placement) and classification stages, we reveal that scallops consistently under-perform due to intrinsic visual properties that make them difficult to distinguish from seafloor backgrounds and other species.

Our key findings include: (1) Performance gaps persist under balanced training conditions, indicating intrinsic rather than data-driven challenges; (2) Class interdependence effects show that models learn target identity partly through negative examples from other species; (3) For imbalanced data, the training distribution creates fundamental precision-recall tradeoffs that should be aligned with operational requirements; and (4) Background discrimination is the primary bottleneck in underwater detection.

These insights lead to practical recommendations: applications prioritizing recall, such as conservation monitoring, should use balanced training, while commercial operations needing high precision should prefer imbalanced training. Since background separation is the main challenge, research should focus on improving localization over classification, using algorithmic advances or additional sensing. This systematic analysis aims to foster targeted solutions for underwater object detection.

{\small
\bibliographystyle{ieee_fullname}
\bibliography{references}
}

\end{document}


\title{Are All Marine Species Created Equal?\\
Performance Disparities in Underwater Object Detection \\ 
\vspace{\baselineskip}
\large{Supplementary Material}}

\renewcommand{\figurename}{Supplementary Figure}

\maketitle
\section*{Overview}

Here we provide additional implementation details for our systematic analysis (Section~\ref{sec:implement}); extended results, mainly on the RUOD dataset~\cite{fu2023rethinking}; and perform ablations for the architectures used in our systematic analysis. Section~\ref{sec:results} supplements the results from the main paper, while Section~\ref{sec:architecture} provides information on the performance of alternative architectures to support the generality of our results.

\section{Additional Implementation Details}
\label{sec:implement}

\subsection{Localization}

The model was loaded and trained with the default parameters, \ie 30 epochs, with images resized to 640x640 pixels and data augmentations during training, including hue, saturation, brightness, translation, image scaling, horizontal flip and mosaic augmentation. The batch size was set to 8. All localization experiments were run on an NVIDIA H100 GPU.

\subsection{Classification}

For the classification experiments, all images are resized to 224x224 pixels and normalized using ImageNet means and standard deviation.  We fine-tune this model for 30 epochs with a batch size of 32. The training images are augmented with random horizontal flip and random rotation within $\pm$15 degrees. We use standard cross-entropy loss and Adam as optimizer with an initial learning rate of 0.001 that we reduce by the factor 0.5 every time no validation improvement has been recorded for 3 epochs, stopping at the minimum learning rate of 0.00003. All classification experiments were conducted on an Apple MacBook Pro M4 with MPS.

\section{Additional Results Analysis}
\label{sec:results}
We highlight in our paper that the main focus lies on the DUO dataset, as its extremely strong class-imbalance allows for effective dataset ablations and highly indicative results. The same experiments conducted on a modified RUOD dataset (RUOD-4C) confirm the main findings, though the effects are less pronounced. For this reason, only the major RUOD-4C results are reported in our paper and some additional visuals can be found in Section~\ref{subsec:RUOD}. 
Similarly, when we present the classification results in the main paper (Paper Sec. 5.4), we only report the inter-class dependencies regarding scallop performance, as these are considered the key insights. Section~\ref{subsec:inter-class} provides proof of that significance compared to other classes.

\subsection{RUOD-4C}
\label{subsec:RUOD}
\subsubsection{Additions to the Localization Study}
\begin{figure*}
    \centering
    \begin{subfigure}{0.23\textwidth}
        \includegraphics[width=\linewidth]{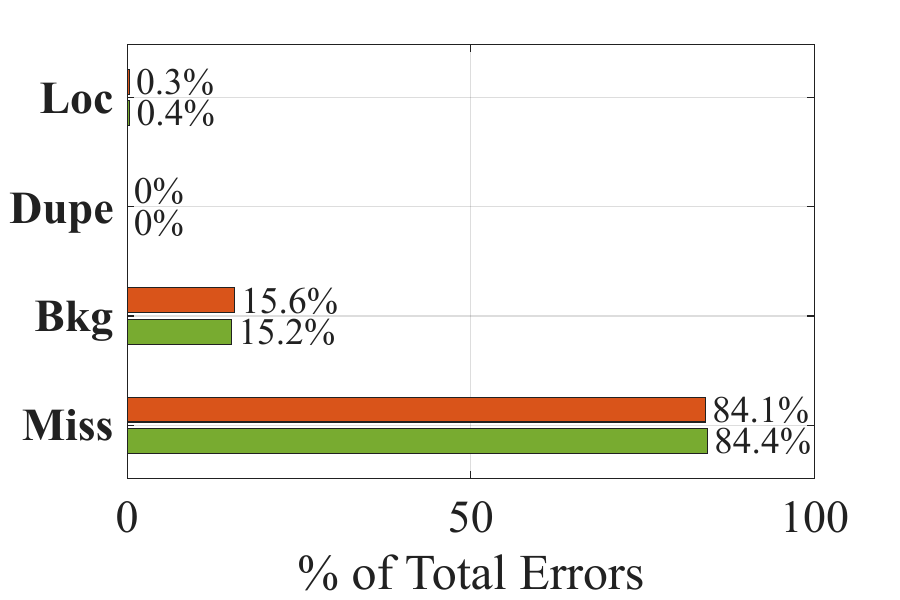}
        \caption{Holothurian }
    \end{subfigure}
    \hfill
    \begin{subfigure}{0.23\textwidth}
        \includegraphics[width=\linewidth]{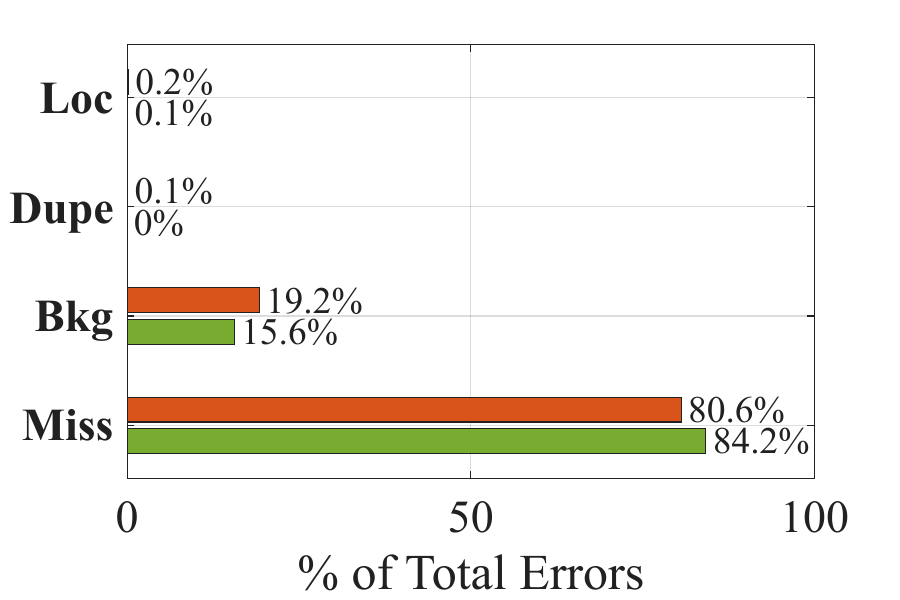}
        \caption{Echinus }
    \end{subfigure}
    \hfill
    \begin{subfigure}{0.23\textwidth}
        \includegraphics[width=\linewidth]{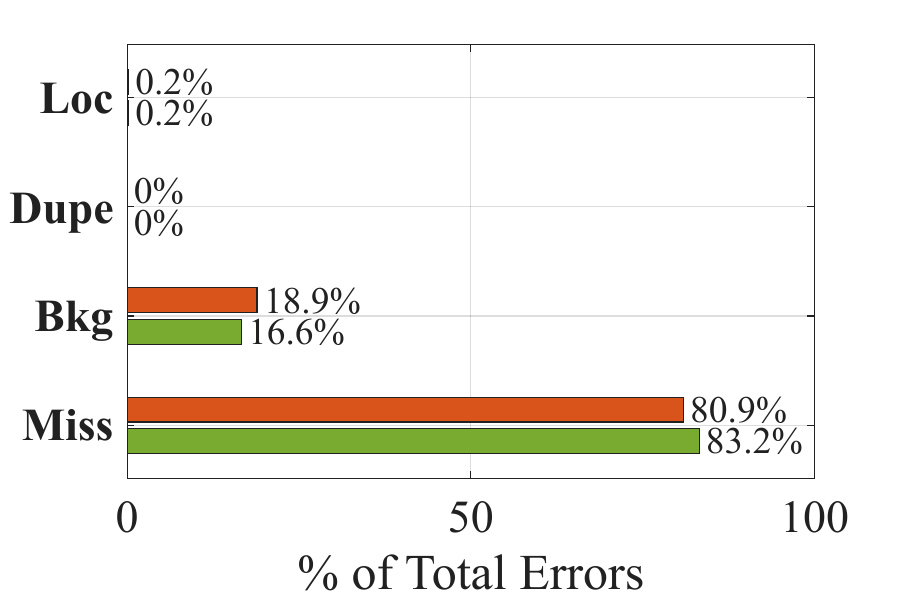}
        \caption{Starfish }
    \end{subfigure}
    \hfill
    \begin{subfigure}{0.23\textwidth}
        \includegraphics[width=\linewidth]{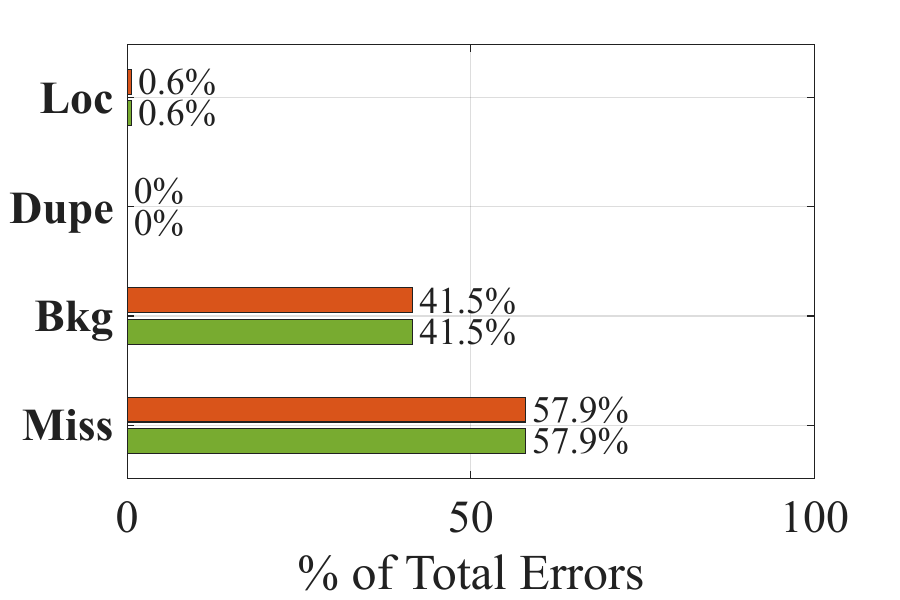}
        \caption{Scallop}
    \end{subfigure}

    \caption{Distribution of TIDE error types across all classes for both original (red) and balanced (green) RUOD-4C datasets. The error profiles do not change much from original to balanced. Missed ground truth objects are dominant type, followed by background error in every class.}
    \label{fig:tide_error_breakdown_type}
    \vspace*{-0.2cm}
\end{figure*}

The TIDE tool has contributed significantly to our localization study and its results on DUO were described in detail in Sec. 4.4 and Sec. 4.5 of the main paper. We apply TIDE to RUOD-4C (see Figure~\ref{fig:tide_error_breakdown_type}) and predominantly observe the error type distribution that we identified for the balanced DUO data,~\ie~objects that blend into the background are completely overlooked and pose the biggest error risk. This pattern is expected to be evident in both, the original and balanced version of RUOD-4C, because its original class distribution is already nearly balanced ( holothurian: 22\%, echinus: 33\%, scallop: 22\%, starfish: 23\%). Hence, these results support the claims made in our main paper.

Additionally, in light of the bounding box accuracy, this error profile with negligible localization errors is consistent with the findings on DUO and the IoU threshold comparison, indicating high box accuracy and minimal class differences in determining the exact object position.

\subsubsection{Additions to the Classification Study}
\begin{figure}[t]
  \centering
  \includegraphics[width=0.9\linewidth]{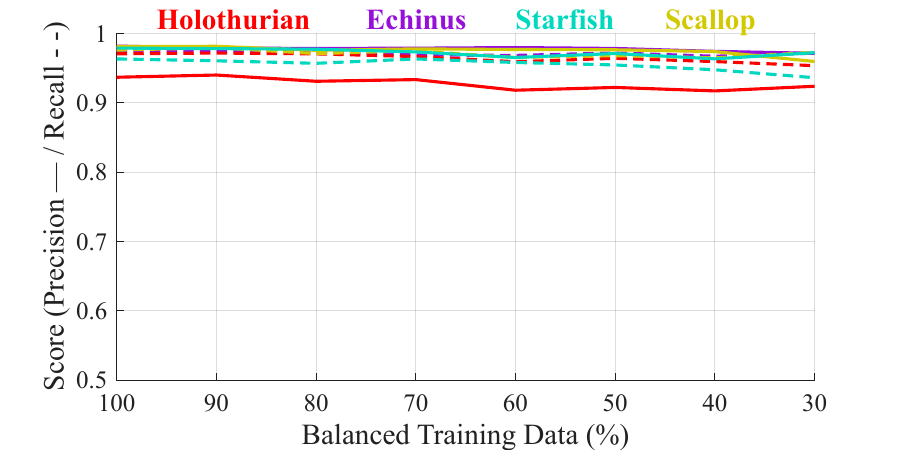}
  \caption{Per class classification performance on incrementally reduced balanced RUOD-4C subsets. Both, precision (solid lines) and recall (dashed lines) are generally high for all classes without major discrepancies. Precision values are slightly more variable than recall.}
  \label{fig:cls_balanced_p_r}
  \vspace*{-0.25cm}
\end{figure}

One of the major insights from our classification study is the precision-recall tradeoff between balanced and imbalanced data distributions. However, as already explained in our paper, the RUOD-4C dataset does not notably show that because of its original fairly even distribution, leading to no significantly different subsets to compare. Nonetheless, we examine the balanced RUOD-4C data in more detail, as we have done for DUO in the main paper. Figure~\ref{fig:cls_balanced_p_r} shows the consistently high recall across all classes and a similar level for precision but with a larger range between classes. The values spread slightly more with less data in general, resulting in starfish still having better precision than recall, but for all other classes their respective recall is higher. This is coherent with our paper findings: extremely good recall is achieved with balanced data, whereas precision is more variable across classes with the same training data quantity.

\subsection{Inter-Class Dependencies}
\label{subsec:inter-class}
\begin{figure}[t]
  \centering

  \includegraphics[width=\linewidth, trim=80 5 110 0, clip]{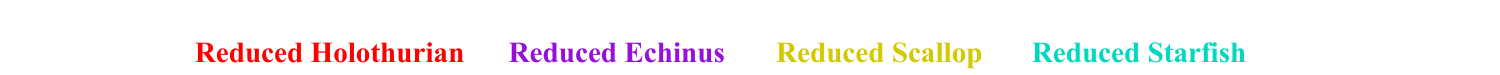}

  \begin{subfigure}[t]{0.49\linewidth}
    \centering
    \includegraphics[width=\linewidth]{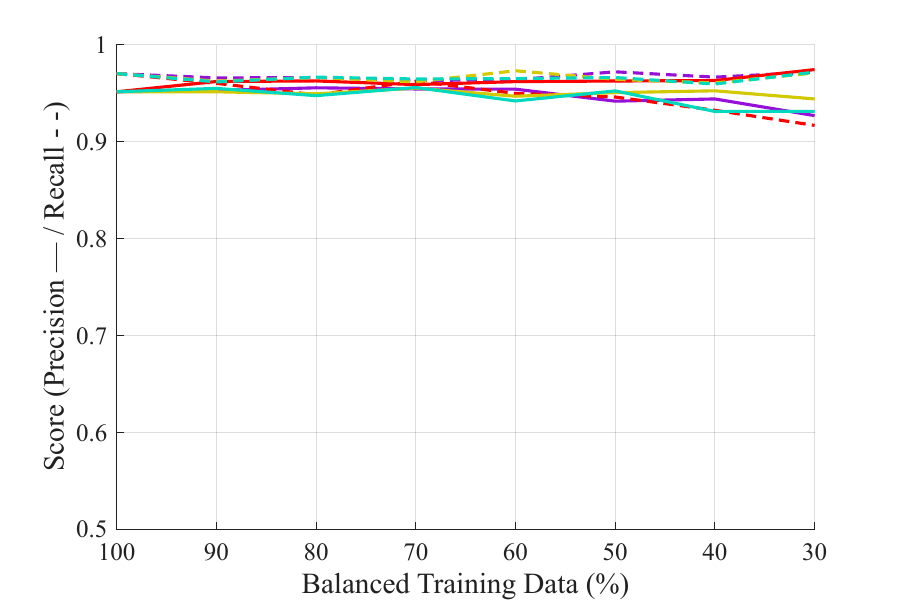}
    \caption{Holothurian Performance}
    \label{fig:ho}
  \end{subfigure}%
  \hfill
  \begin{subfigure}[t]{0.49\linewidth}
    \centering
    \includegraphics[width=\linewidth]{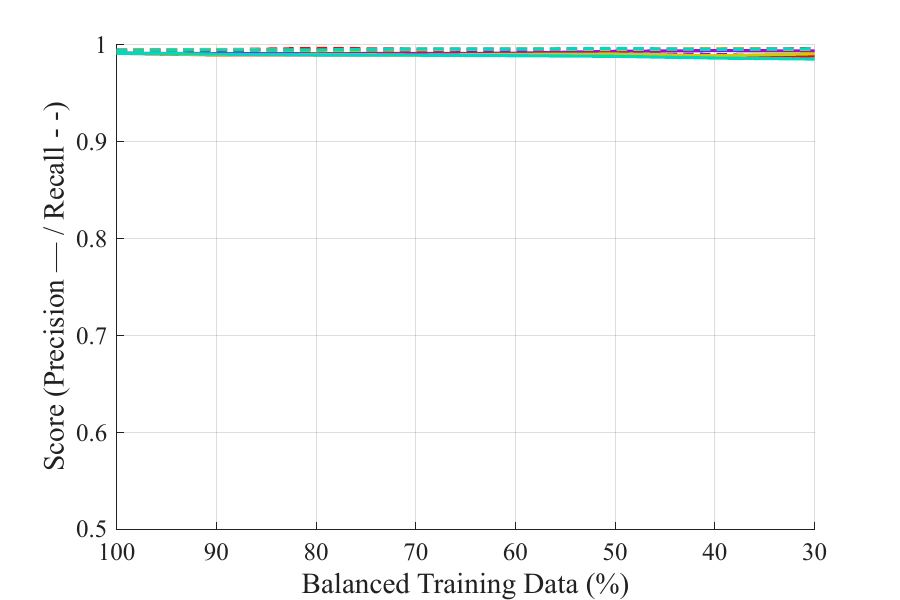}
    \caption{Echinus Performance}
    \label{fig:ec}
  \end{subfigure}
  \hfill
  \begin{subfigure}[t]{0.49\linewidth}
    \centering
    \includegraphics[width=\linewidth]{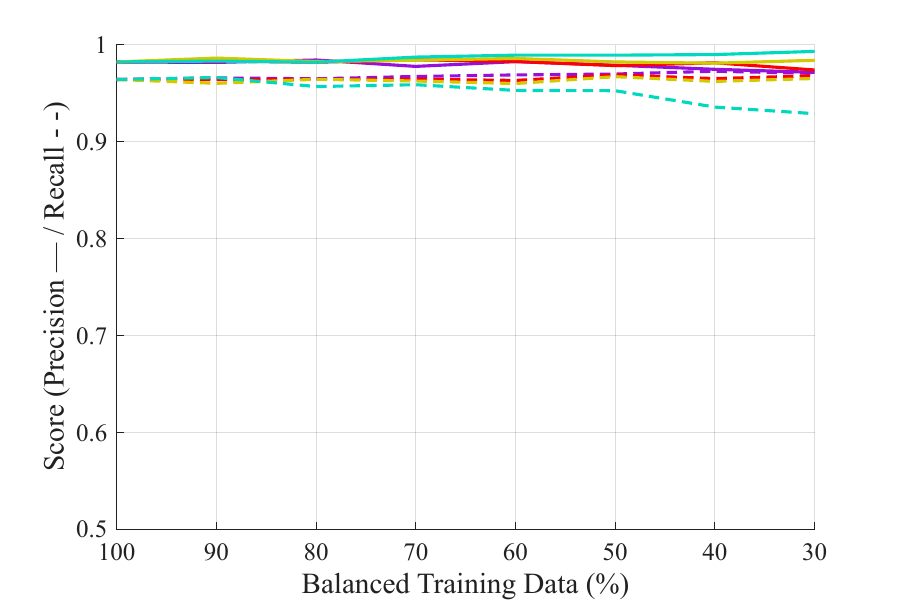}
    \caption{Starfish Performance}
    \label{fig:st}
  \end{subfigure}%
  \hfill
  \begin{subfigure}[t]{0.49\linewidth}
    \centering
    \includegraphics[width=\linewidth]{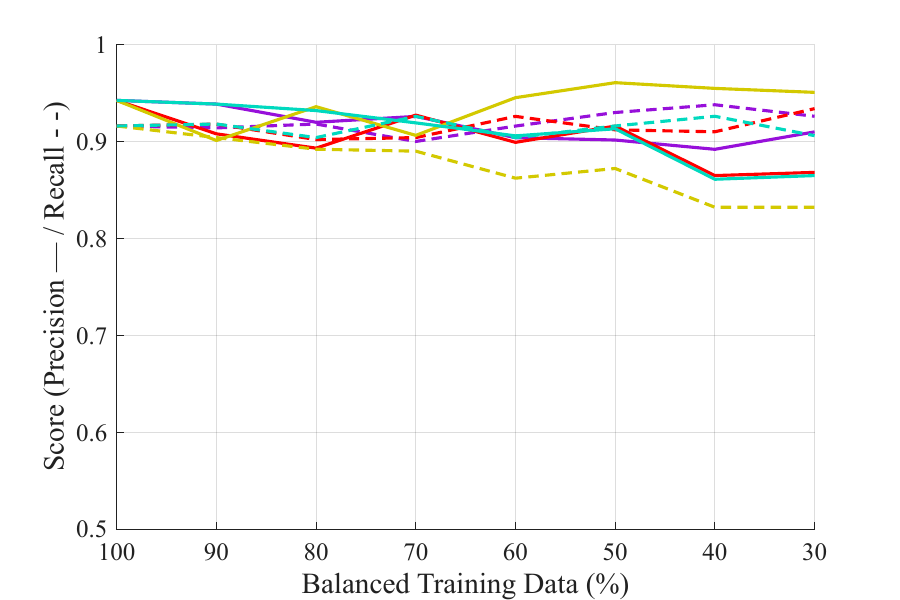}
    \caption{Scallop Performance}
    \label{fig:sc}
  \end{subfigure}

  \caption{Classification performance change of every species when subject to class-specific data reductions in DUO. Echinus is extremely robust, starfish and holothurian are mainly influenced by reductions in their own training data, scallop is sensitive to data decrease in any class and notably shows the largest fluctuations.}
  \label{fig:sensitivity}
  \vspace*{-0.25cm}
\end{figure}

When discussing the classification results in our main paper, we evaluate how sensitive scallops are to changes in data quantity per class (Paper Figure 7). We examine the increase or decrease of scallop precision and recall values from the original reference set to when one class is reduced to only 30\% of its available training instances. This analysis is limited to the scallop class because we found that scallops are significantly more affected by data variations than the other classes, as directly visible in Figure~\ref{fig:sensitivity}. When monitoring performance over gradually decreasing data (one class at a time), the metrics generally changed most with the least data, as expected. However, there are fundamental differences as to what extent. The changes in echinus performance are minimal and fully negligible, whereas some effects on starfish and holothurian metrics can be observed, mostly linked to their own class-specific data. Scallops, on the other hand, rely on all the available data of any class, which is evident in very strong fluctuations and performance changes. This is the key insight of this sub-experiment that we aimed to present in our main paper using Figure 7.

\section{Architecture Ablations}
\label{sec:architecture}
Our experimental framework presented in the main paper includes the systematic decomposition of the object detection pipeline into localization (Paper Sec.~4) and classification (Paper Sec.~5). While we report all representative results based on YOLO11 and ResNet-18 models, we here include complementary analyses with additional architectures. These experiments follow exactly the same framework as in the main paper and are intended to confirm the generality of our findings.

\subsection{Extending the Localization Study with SSD}
\subsubsection{Implementation}
We repeat our localization experiments with the popular Single Shot multibox Detector~\cite{liu2016ssd}, specifically PyTorch's SSD300-VGG16, that is with input size 300x300 and VGG16 as backbone. Apart from the architectural change, we stay consistent with the settings~\ie~30 epochs, a batch size of 8, a confidence threshold of 0.25, a COCO-pretrained model state and a NVIDIA H100 GPU.

\subsubsection{Results}
\begin{figure}[t]
  \centering

  \includegraphics[width=\linewidth, trim=100 5 100 0, clip]{Illustrations/class_label.pdf}

  \begin{subfigure}[t]{0.49\linewidth}
    \centering
    \includegraphics[width=\linewidth]{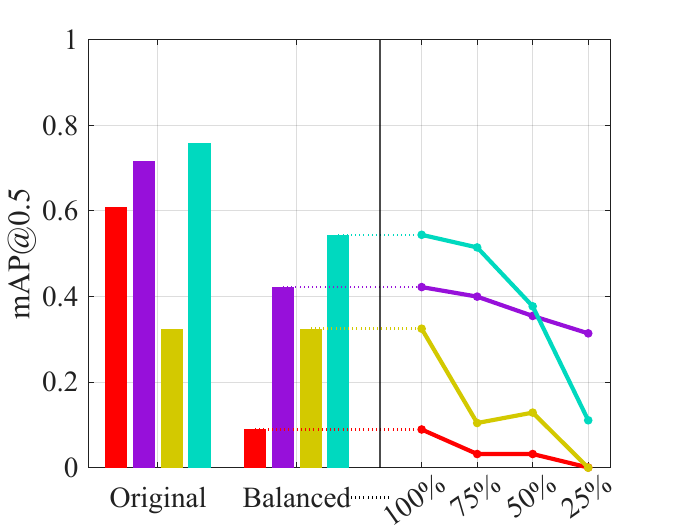}
    \caption{DUO}
    \label{fig:duo_map}
  \end{subfigure}%
  \hfill
  \begin{subfigure}[t]{0.49\linewidth}
    \centering
    \includegraphics[width=\linewidth]{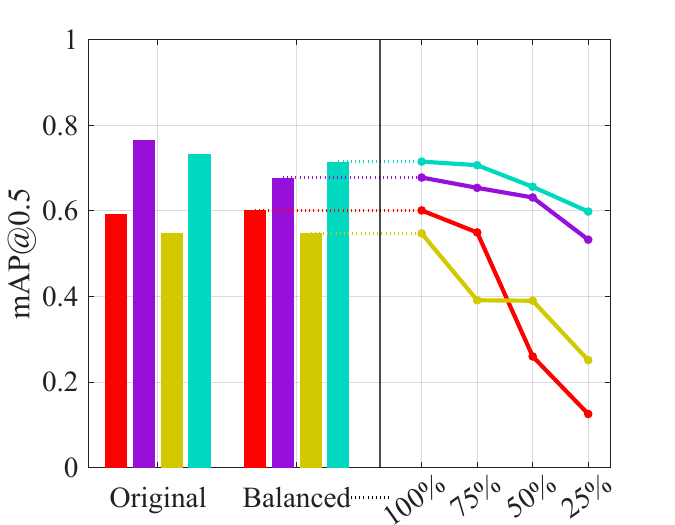}
    \caption{RUOD-4C}
    \label{fig:ruod-4c_map}
  \end{subfigure}
  
  \caption{SSD localization performance per class compared across datasets. Bar charts indicate mAP@0.5 for the originally imbalanced and balanced data, line charts represent gradually reduced sets. Class differences are evident in all setups and SSD struggles with holothurians more than YOLO11. Still, the scallop class is mostly the worst-performing.}
  \label{fig:loc_mAP_full_balanced_reduced_SSD}
  \vspace*{-0.25cm}
\end{figure}

Our main conclusions from the localization study as presented in our paper include that performance gaps remain evident under balanced conditions, pointing at inherent species characteristics and that foreground-background separation is the most challenging object detection step. In that regard, we show that once the presence of a target is detected, there are no significant differences between classes when determining the exact position afterwards anymore. These findings are supported by the results obtained using the SSD detector instead of YOLO11 in our experiments: Figure~\ref{fig:loc_mAP_full_balanced_reduced_SSD} and Table~\ref{tab:loc_TP_FP_FN_SSD} compare various performance metrics between the different distributions and confirm persisting class-disparities. The SSD model achieves notably lower mAPs on every class in every dataset than we obtain with YOLO11 in our main paper, yet relative performance trends are similar. Although SSD is significantly challenged by holothurian mAP and echinus recall in the balanced DUO, the scallop class can still be identified as the most problematic one overall. It usually has the highest error rates and lowest recall and mAP.

\definecolor{blue}{RGB}{70, 130, 180} 

\newcommand{\colormap}[2]{%
  \colorbox{blue!#1!white}{#2}%
}

\begin{table}
\centering
\setlength{\tabcolsep}{1pt}
\renewcommand{\arraystretch}{1.4}
\scriptsize
\begin{tabular}{>{\raggedright\arraybackslash}p{0.5cm} 
                >{\centering\arraybackslash}p{1.215cm}
                >{\centering\arraybackslash}p{1.215cm}
                >{\centering\arraybackslash}p{1.215cm}
                >{\centering\arraybackslash}p{1.215cm}
                >{\centering\arraybackslash}p{1.215cm}
                >{\centering\arraybackslash}p{1.215cm}}
\toprule
\ & \multicolumn{3}{c}{\textbf{Original Distribution (Imbalanced)}} & \multicolumn{3}{c}{\textbf{Balanced Sets}} \\
\cmidrule(lr){2-4} \cmidrule(lr){5-7}
 & TPR & FDR & FNR & TPR & FDR & FNR \\
\midrule
Ho & 
\colormap{66.4}{66.4}\,\textbf{/}\,\colormap{75.6}{75.6} & \colormap{21.9}{21.9}\,\textbf{/}\,\colormap{23.1}{23.1} & \colormap{33.6}{33.6}\,\textbf{/}\,\colormap{24.4}{24.4} & 
\colormap{63.4}{63.4}\,\textbf{/}\,\colormap{77.6}{77.6} & \colormap{66.1}{66.1}\,\textbf{/}\,\colormap{25.2}{25.2} & \colormap{36.6}{36.6}\,\textbf{/}\,\colormap{22.4}{22.4} \\

Ec & 
\colormap{77.1}{77.1}\,\textbf{/}\,\colormap{83.7}{83.7} & \colormap{13.3}{13.3}\,\textbf{/}\,\colormap{12.5}{12.5} & \colormap{22.9}{22.9}\,\textbf{/}\,\colormap{16.3}{16.3} & 
\colormap{43.8}{43.8}\,\textbf{/}\,\colormap{73.6}{73.6} & \colormap{1.9}{1.9}\,\textbf{/}\,\colormap{7.0}{7.0} & \colormap{56.2}{56.2}\,\textbf{/}\,\colormap{26.5}{26.5} \\

St & 
\colormap{80.0}{80.0}\,\textbf{/}\,\colormap{78.5}{78.5} & \colormap{13.4}{13.4}\,\textbf{/}\,\colormap{11.3}{11.3} & \colormap{20.0}{20.0}\,\textbf{/}\,\colormap{21.5}{21.5} & 
\colormap{63.5}{63.5}\,\textbf{/}\,\colormap{76.8}{76.8} & \colormap{21.1}{21.1}\,\textbf{/}\,\colormap{10.0}{10.0} & \colormap{36.5}{36.5}\,\textbf{/}\,\colormap{23.2}{23.2} \\

Sc & 
\colormap{49.8}{49.8}\,\textbf{/}\,\colormap{69.1}{69.1} & \colormap{63.9}{63.9}\,\textbf{/}\,\colormap{30.2}{30.2} & \colormap{50.2}{50.2}\,\textbf{/}\,\colormap{30.9}{30.9} & 
\colormap{49.8}{49.8}\,\textbf{/}\,\colormap{69.1}{69.1} & \colormap{63.9}{63.9}\,\textbf{/}\,\colormap{30.2}{30.2} & \colormap{50.2}{50.2}\,\textbf{/}\,\colormap{30.9}{30.9} \\

\bottomrule
\end{tabular}
 \caption{SSD localization performance rates (True Positive Rate (TPR), False Discovery Rate (FDR), and False Negative Rate (FNR)) per class for DUO\,/\,RUOD-4C. Scallop performs better in SSD than in YOLO11 in terms of TPR/FNR but worse in FDR. It usually remains the worst-performing class.}
 \label{tab:loc_TP_FP_FN_SSD}
\end{table}

The experiments using SSD also further verify that drawing accurate bounding boxes is not significantly easier or more difficult for certain classes, once the object has already been detected. Table~\ref{tab:map_deviation_SSD} shows that class disparities become smaller when focusing on better-aligned boxes in mAP@0.5:0.95, consistent with the results in our main paper (Paper Table 3).

\begin{table}
\centering
  \setlength{\tabcolsep}{4.5pt}
  \scriptsize
  \begin{tabular}{lcccc}
    \toprule
    & \multicolumn{2}{c}{\textbf{mAP@0.5}} 
    & \multicolumn{2}{c}{\textbf{mAP@0.5:0.95}} \\
    \cmidrule(lr){2-3} \cmidrule(lr){4-5}
    \textbf{Class} & \textbf{Value} & \textbf{Deviation} 
                   & \textbf{Value} & \textbf{Deviation} \\
    \midrule
    Holothurian & 
    $0.09$\,\textbf{/}\,$0.60$ &  
    $-0.45$\,\textbf{/}\,$-0.11$ & 
    $0.03$\,\textbf{/}\,$0.30$ &  
    $-0.26$\,\textbf{/}\,$-0.09$ \\
    
    Echinus     & 
    $0.42$\,\textbf{/}\,$0.68$ &  
    $-0.12$\,\textbf{/}\,$-0.03$ & 
    $0.30$\,\textbf{/}\,$0.33$ &  
    $0.00$\,\textbf{/}\,$-0.06$ \\
    
    Starfish    & 
    $0.54$\,\textbf{/}\,$0.71$ &  
    $0.00$\,\textbf{/}\,$0.00$ & 
    $0.30$\,\textbf{/}\,$0.39$ &  
    $0.00$\,\textbf{/}\,$0.00$ \\
    
    Scallop     & 
    $0.32$\,\textbf{/}\,$0.55$ &  
    $-0.22$\,\textbf{/}\,$-0.16$ & 
    $0.19$\,\textbf{/}\,$0.29$ &  
    $-0.11$\,\textbf{/}\,$-0.10$ \\
    
    \bottomrule
  \end{tabular}
   \caption{DUO\,/\,RUOD-4C per-class mAP results of the balanced dataset and the deviation from the best‐performing class using SSD model. The gaps decrease with increasing IoU threshold confirming the findings of our main paper.
   }
   \label{tab:map_deviation_SSD}
  \vspace*{-0.2cm}
\end{table}

\subsection{Extending the Classification Study with MobileNet and Vision Transformer}
\subsubsection{Implementation}
To verify our classification results we conduct the same experiments as in Paper Sec. 5 again with the lightweight CNN architecture MobileNetV2~\cite{sandler2018mobilenetv2} and the transformer-based VisionTransformer-B/16~\cite{dosovitskiy2020image} as classifiers. Both are implemented in the exact same way as ResNet-18 in PyTorch, initialized with pretrained weights and then fine-tuned for 30 epochs, using a batch size of 32, input image size of 224x224, standard cross-entropy loss and the Adam optimizer.

\subsubsection{Results}
\begin{figure}[t]
  \centering

  \includegraphics[width=\linewidth, trim=100 5 100 0, clip]{Illustrations/class_label.pdf}
  \vspace*{-0.3cm}

  \begin{minipage}{\linewidth}
    \centering
    \begin{subfigure}[t]{0.48\linewidth}
      \centering
      \includegraphics[width=\linewidth]{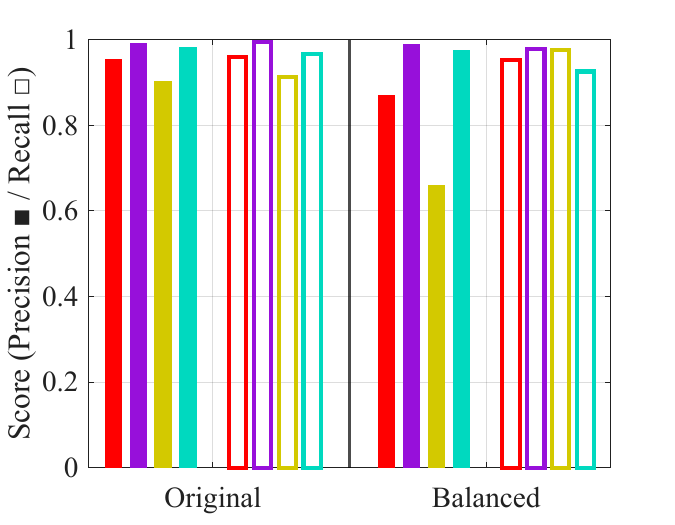}
      \caption{DUO}
      \label{fig:duo_mnv2}
    \end{subfigure}%
    \hfill
    \begin{subfigure}[t]{0.48\linewidth}
      \centering
      \includegraphics[width=\linewidth]{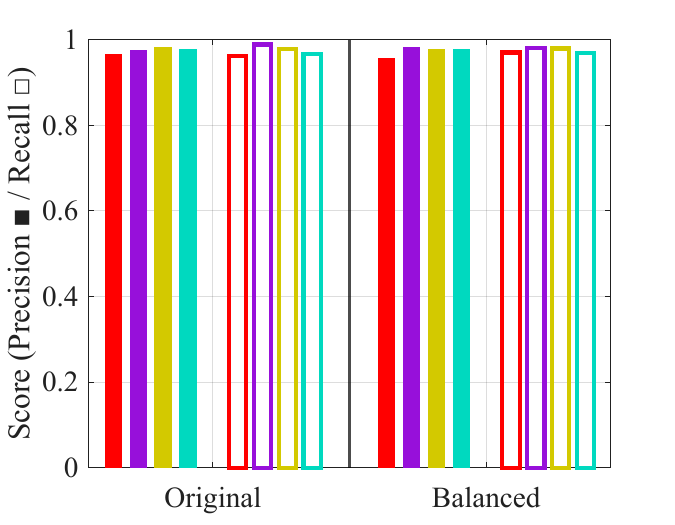}
      \caption{RUOD-4C}
      \label{fig:ruod-4c_mnv2}
    \end{subfigure}
    \footnotesize\textbf{MobileNetV2} \\[0.2cm]
  \end{minipage}

  \vspace{0.3cm} %

  \begin{minipage}{\linewidth}
    \centering
    \begin{subfigure}[t]{0.48\linewidth}
      \centering
      \includegraphics[width=\linewidth]{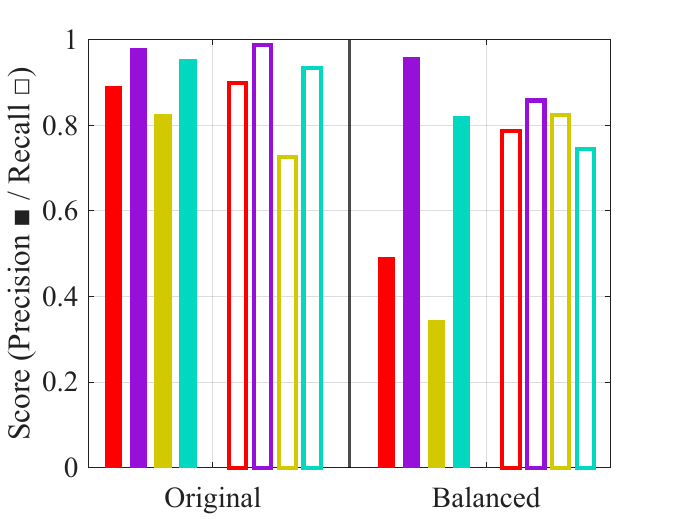}
      \caption{DUO}
      \label{fig:duo_vit}
    \end{subfigure}%
    \hfill
    \begin{subfigure}[t]{0.48\linewidth}
      \centering
      \includegraphics[width=\linewidth]{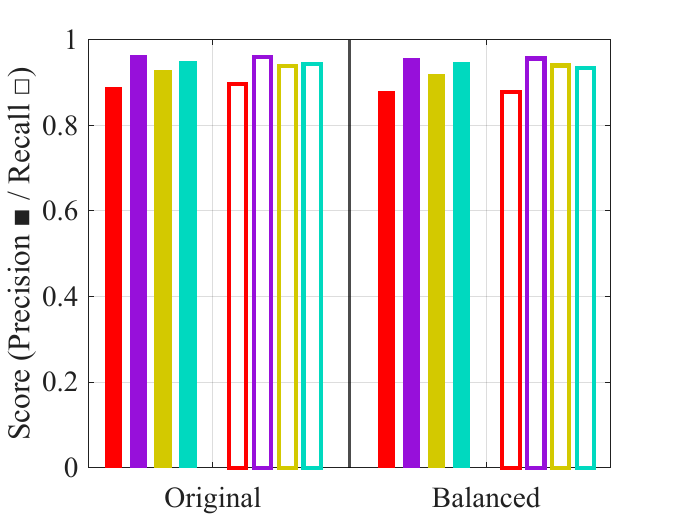}
      \caption{RUOD-4C}
      \label{fig:ruod-4c_vit}
    \end{subfigure}
    \footnotesize \textbf{VisionTransformer} \\[0.2cm]
  \end{minipage}

  \caption{Classification precision and recall for DUO (left column) and RUOD-4C (right column) across the additional architectures MobileNetV2 (top row) and ViT-B/16 (bottom row). While performance is steady on a high level for RUOD-4C, there are notable precision and recall changes with class-distribution in DUO. These patterns are consistent with ResNet-18 results reported in the main paper.}
  \label{fig:cls_arch_ablation}
\end{figure}

The main take-aways from our classification study presented in the paper include that performance is usually very high compared to the localization stage, but there is a notable tradeoff between recall and precision when comparing classification metrics between strongly imbalanced and balanced data versions. This is the case for scallops in DUO with all architectures: From imbalanced to balanced, we report a precision drop of 27.9\% plus a recall gain of 4.6\% using ResNet-18 in the main paper and we can see here in Figure~\ref{fig:cls_arch_ablation} a precision decrease of 24.3\% and recall increase by 6.4\% for MobileNetV2 as well as drastic -48\% in precision and +9.8\% in recall for VisionTransformer. Since the original and balanced RUOD-4C are very similar, there are no significant changes observed with other architectures, but this too stays consistent with the ResNet results.

In our paper, we also highlight the sensitivity of scallops as a minority and visually ambiguous class. Table~\ref{tab:cls-abl-interclass} reports the results from the same experiment (reducing one class only) using MobileNetV2 and VisionTransformer. Both reveal the strong class-interdependence of scallops and underline their reliance on negative examples from other species. While holothurian and starfish also show some major performance changes with VisionTransformer, they stay mostly sensitive to their own training samples. All data reduction effects are pronounced significantly stronger in the ViT model, which might be related to transformer architectures generally requiring much larger datasets than CNNs for better performance~\cite{dosovitskiy2020image}.

\definecolor{myblue}{RGB}{20, 100, 160}

\definecolor{mygreen}{RGB}{0, 180, 0} %

\newcommand{\colornegative}[2]{%
  \colorbox{myblue!#1!white}{\strut #2}%
}

\newcommand{\colorpositive}[2]{%
  \colorbox{mygreen!#1!white}{\strut #2}%
  }
  
\begin{table}
\centering
\scriptsize
\setlength{\tabcolsep}{3pt} %
\begin{tabular}{l@{\hskip 2pt}cccc@{\hskip 2pt}cccc}
\toprule
 & \multicolumn{4}{c}{\textbf{MobileNetV2}} & \multicolumn{4}{c}{\textbf{ViT-B/16}} \\
\cmidrule(lr){2-5} \cmidrule(lr){6-9}
 Reduced: & Ec & Ho & Sc & St & Ec & Ho & Sc & St \\
\midrule
Ec -- P: & \colorpositive{0.4}{0.2\%} & \colornegative{0.4}{-0.2\%} & \colornegative{0.6}{-0.3\%} & \colornegative{1.0}{-0.5\%} 
   & \colorpositive{1.4}{0.7\%} & \colornegative{5.8}{-2.9\%} & \colornegative{1.6}{-0.8\%} & \colornegative{2.6}{-1.3\%} \\
\textcolor{white}{Ec --} R: & \colornegative{0.8}{-0.4\%} & \colorpositive{0.0}{0.0\%} & \colorpositive{0.2}{0.1\%} & \colorpositive{0.4}{0.2\%} 
   & \colornegative{6.6}{-3.3\%} & \colornegative{1.0}{-0.5\%} & \colornegative{1.0}{-0.5\%} & \colorpositive{0.0}{0.0\%} \\
\midrule
Ho -- P: & \colornegative{2.6}{-1.3\%} & \colorpositive{3.8}{1.9\%} & \colornegative{1.6}{-0.8\%} & \colornegative{2.6}{-1.3\%} 
   & \colornegative{26.8}{-13.4\%} & \colorpositive{2.6}{1.3\%} & \colornegative{15.6}{-7.8\%} & \colornegative{13.4}{-6.7\%} \\
\textcolor{white}{Ho --} R: & \colorpositive{1.8}{0.9\%} & \colornegative{7.2}{-3.6\%} & \colorpositive{0.6}{0.3\%} & \colorpositive{1.6}{0.8\%} 
   & \colorpositive{5.6}{2.8\%} & \colornegative{63.0}{-31.5\%} & \colornegative{1.8}{-0.9\%} & \colorpositive{2.2}{1.1\%} \\
\midrule
Sc -- P: & \colornegative{5.2}{-2.6\%} & \colornegative{7.2}{-3.6\%} & \colorpositive{3.2}{1.6\%} & \colornegative{10.0}{-5.0\%} 
   & \colornegative{19.2}{-9.6\%} & \colornegative{38.4}{-19.2\%} & \colorpositive{8.4}{4.2\%} & \colornegative{26.8}{-13.4\%} \\
\textcolor{white}{Sc --} R: & \colorpositive{10.4}{5.2\%} & \colorpositive{6.0}{3.0\%} & \colornegative{23.8}{-11.9\%} & \colorpositive{6.0}{3.0\%} 
   & \colornegative{3.8}{-1.9\%} & \colornegative{9.4}{-4.7\%} & \colornegative{78.2}{-39.1\%} & \colorpositive{4.4}{2.2\%} \\
\midrule
St -- P: & \colornegative{0.2}{-0.1\%} & \colornegative{2.0}{-1.0\%} & \colorpositive{0.4}{0.2\%} & \colorpositive{2.4}{1.2\%} 
   & \colornegative{9.2}{-4.6\%} & \colornegative{14.2}{-7.1\%} & \colornegative{2.6}{-1.3\%} & \colorpositive{5.2}{2.6\%} \\
\textcolor{white}{St --} R: & \colorpositive{0.4}{0.2\%} & \colorpositive{0.6}{0.3\%} & \colornegative{1.6}{-0.8\%} & \colornegative{6.4}{-3.2\%} 
   & \colornegative{0.6}{-0.3\%} & \colornegative{3.8}{-1.9\%} & \colornegative{4.0}{-2.0\%} & \colornegative{17.6}{-8.8\%} \\
\bottomrule
\end{tabular}
\caption{Relative change of classification performance metrics in DUO from 100\% of available data to 30\% per class. The reduced class is specified in the header (Ec = Echinus, Ho = Holothurian, Sc = Scallop, St = Starfish). We report positive (green) and negative (blue) changes in Precision (P) and Recall (R). The scallop class as a whole is the one most affected. Inter-class dependencies are generally more visible for VisionTransformer.}
\label{tab:cls-abl-interclass}
\end{table}

Overall the results for the respective datasets remain consistent across alternative architectures and serve as further proof for our conclusions in the main paper.

{\small
\bibliographystyle{ieee_fullname}
\bibliography{Bibliography}
}